\documentclass{article}

\usepackage[final, nonatbib]{nips_2018}

\usepackage[utf8]{inputenc} 
\usepackage[T1]{fontenc}    
\usepackage{hyperref}       
\usepackage{url}            
\usepackage{booktabs}       
\usepackage{amsfonts}       
\usepackage{nicefrac}       
\usepackage{microtype}      
\usepackage{amsmath}
\usepackage{amssymb}
\usepackage{helvet}
\usepackage{courier}
\usepackage{graphicx}
\usepackage{booktabs}
\usepackage{xcolor}
\usepackage{multirow}
\usepackage{enumerate}
\usepackage[numbers]{natbib}

\PassOptionsToPackage{options}{natbib}

\begin{document}

\title{ Evaluating Generative Adversarial Networks on Explicitly Parameterized Distributions }
\author{Shayne O'Brien, Matt Groh, Abhimanyu Dubey\\
\{\texttt{shayneob}, \texttt{groh}, \texttt{dubeya}\}@mit.edu \\
Massachusetts Institute of Technology\\
}
\maketitle
\begin{abstract}
\begin{quote}
The true distribution parameterizations of commonly used image datasets are inaccessible. Rather than designing metrics for feature spaces with unknown characteristics, we propose to measure GAN performance by evaluating on explicitly parameterized, synthetic data distributions. As a case study, we examine the performance of 16 GAN variants on six multivariate distributions of varying dimensionalities and training set sizes. In this learning environment, we observe that: GANs exhibit similar performance trends across dimensionalities; learning depends on the underlying distribution and its complexity;  the number of training samples can have a large impact on performance; evaluation and relative comparisons are metric-dependent; diverse sets of hyperparameters can produce a ``best'' result; and some GANs are more robust to hyperparameter changes than others. These observations both corroborate findings of previous GAN evaluation studies and make novel contributions regarding the relationship between size, complexity, and GAN performance.
\end{quote}
\end{abstract}

\section{Introduction}



Generative adversarial network (GAN) optimization stability and convergence properties remain poorly understood despite the introduction of hundreds of GAN variants since their conception \cite{goodfellow2014generative,theganzoo}. While GAN learning and performance behavior has been studied~\cite{heusel2017gans,lucic2017gans,salimans2016improved}, most existing work examining this relationship focuses on image datasets for which the underlying distribution parameterization is inaccessible~\cite{arjovsky2017wasserstein,arora2017generalization, borji2018pros,lee2017ability,theis}. This is problematic since claims of behavior that are made by modeling an unknown target distribution require a strong assumption for generalizability.

The goal of generative modeling is to approximate a distribution $p_{d}$ by learning a parameterized distribution $p_{g}$, where both $p_{d}$ and $p_{g}$ are defined over samples. If we do not have full access to $p_{d}$, generalizability requires us to assume that the modeled dataset is a reasonable proxy for the family of distributions from which it was sampled. Without this assumption that is often only implicitly made, using images to understand GAN behavior limits conclusions to the data context being modeled.

We seek to address a gap in the literature by investigating GAN variant performance on datasets for which we have full access to the distribution parameterization. This allows us to study empirical performance on data where we can make claims of model behavior that generalize to the full distribution, as opposed to on image datasets for which this is not necessarily true. To this end, we examine the performance of 16 GAN variants on six explicitly parameterized multivariate distributions of four different dimensionalities and three different training set sizes. 

Across 20 grid search trials, we observe that: (1) GANs exhibit similar performance trends across dimensionalities, (2) learning depends on the underlying distribution and its complexity, (3) the number of training samples can have a large impact on performance, (4) evaluation and relative comparisons are metric-dependent, (5) diverse sets of hyperparameters can produce a ``best'' result, and (6) some GANs are more robust to hyperparameter changes than others. These findings corroborate those of previous GAN evaluation studies as well as contribute novel insights regarding the relationship between size, complexity, and GAN performance.\footnote{All code is publicly available at \url{https://github.com/shayneobrien/explicit-gan-eval}.}

\section{Related Work}

One notable work in this area by \citet{lucic2017gans} compares seven GAN variants in terms of modeling ability and optimization stability. The authors find that as computational budget increases, all tested models reach similar Frech\'et Inception Distance on the MNIST, Fashion-MNIST, CIFAR10, and CelebA datasets; and F1, precision, and recall on a synthetic dataset of convex polygons. They also discuss the difficulties of comparing GANs due to multiple valid ways to analyze performance. 

\citet{santurkar} measure Inception Score and classification accuracy and report that the five GAN variants they train do not succeed at capturing distributional properties of the training set on the CelebA and LSUN datasets. The authors observe that the GAN distributions exhibit significantly less diversity at test time compared to the evaluation dataset, suggesting $p_g$ is far from $p_d$.

In another study, \citet{divergencetraining} evaluate GAN variant performance based on the original GAN criterion, least squares, maximum mean discrepancy, and improved Wasserstein distance. They show that for the three GAN variants they consider, test-time metrics do not favor networks that use the same training-time criterion on the MNIST, CIFAR10, LSUN, and Fashion-MNIST image datasets. The authors also examine performance as a function of sample size and show that some GANs exhibit faster performance increases than others as the number of training samples increases.


Lastly, \citet{borji2018pros} provide a thorough discussion of the strengths and weaknesses of 26 quantitative and qualitative measures used for evaluating GANs trained on image datasets. They  conclude that there is no single, best GAN evaluation measure. The authors suggest benchmarking models under identical architectures and computational budgets, and using more than a single metric to make comparisons.


\section{Experimental Setup}
In GANs, we define a prior probability distribution on input noise variables $p_{z}(\mathbf{z})$ and represent a mapping to the target data space $p_{d}(\mathbf{x})$ as $G(\mathbf{z}, \theta_{G})$, where $G$ is a fully differentiable neural network called the \textit{generator} and $\theta_{G}$ are its parameters. We train $G$ by simultaneously learning a fully differentiable network $D$, called the \textit{discriminator} or \textit{critic} and defined by $D(\mathbf{x}, \theta_{D})$, that helps $G$ during training. Whereas $G$ is trained to mimic $p_{d}$, the learning objective, output, and precise task of $D$ vary depending upon the GAN variant.
\subsection*{Models}

As a case study, we examine the same seven GAN variants evaluated by \citet{lucic2017gans} and nine additional GAN variants that have been popularly discussed since their study was published. The primary difference between considered variants is whether the discriminator output can be interpreted as a probability (MMGAN, NSGAN \cite{goodfellow2014generative}, RaGAN \cite{ragan}, DRAGAN \cite{kodali2017dragan}, FisherGAN \cite{fishergan}, InfoGAN \cite{infogan}, ForwGAN, RevGAN, HellingerGAN, PearsonGAN, JSGAN \cite{fgan})
or is unbounded (WGAN \cite{arjovsky2017wasserstein}, WGANGP \cite{gulrakani2017wassersteingp}, LSGAN \cite{lsgan}, BEGAN \cite{berthelot2017began}).
We summarize these models in Table~\ref{tab:gansfncs}.

\begin{table*}[h]
  \centering
  \resizebox{\textwidth}{!}{%
  \begin{tabular}{l l} \\ \toprule
    GAN Variant Loss Functions &  \\ \toprule

    $\mathcal{L}^{\textnormal{MMGAN}}$ = $\mathbb{E}$[$\log(D(\mathbf{x}))] +  \mathbb{E}$[$\log(1-D(G(\mathbf{z})))]$ 
    & 
    $\mathcal{L}^{\textnormal{RaGAN}}$ =  $\mathbb{E}$[$\log( D(\mathbf{x}) - D(G(\mathbf{z})))] +  \mathbb{E}[\log(1-(D(G(\mathbf{z})) - D(\mathbf{x})))]$  \\ \midrule
    
    $\mathcal{L}^{\textnormal{NSGAN}}$ = $\mathbb{E}$[$\log(D(\mathbf{x}))] - \mathbb{E}$[$\log(D(G(\mathbf{z})))]$
    & 
    $\mathcal{L}^{\textnormal{LSGAN}}$ = $-\mathbb{E}[(D(\mathbf{x}) - 1)^{2}] + \mathbb{E}[D(G(\mathbf{z}))^{2}]$ \\ \midrule

    $\mathcal{L}^{\textnormal{WGAN}}$ = $-\mathbb{E}$[$D(\mathbf{x})] + \mathbb{E}$[$D(G(\mathbf{z}))]$
    & 
    $\mathcal{L}^{\textnormal{BEGAN}}$ = $\mathbb{E} [ \|\mathbf{x} - D_{\textnormal{AE}}(\mathbf{x}) \|_{1}] - k_{t}\mathbb{E}[\| G(\mathbf{z}) - D_{\textnormal{AE}}(G(\mathbf{z})) \|_{1}]$  \\ \midrule
    
    $\mathcal{L}^{\textnormal{WGANGP}}$ = $\mathcal{L}^{\textnormal{WGAN}} + \lambda \mathbb{E}$ [($\| \nabla_{\mathbf{z}} D(G(\mathbf{z})) \|_{2} - 1)^{2}]$
    & 
    $\mathcal{L}^{\textnormal{DRAGAN}}$ = $\mathcal{L}^{\textnormal{MMGAN}} + \lambda \mathbb{E}[(\| \nabla_{\mathbf{x}} D(\mathbf{x} + \delta)) \|_{2} - 1)^{2}]]$ \\ \midrule

    $\mathcal{L}^{\textnormal{FisherGAN}}$ = $\mathcal{L}^{\textnormal{WGAN}} + \lambda (1 - \hat{\Omega}(D, G)) - \frac{\rho}{2}(\hat{\Omega}(D, G) - 1)$ 
    & 
    $\mathcal{L}^{\textnormal{InfoGAN}}$ = $\mathcal{L}^{\textnormal{MMGAN}}$ - $\lambda(\mathbb{E}[\log (Q(\mathbf{c}'|\mathbf{x}))])$  \\ \midrule

    $\mathcal{L}^{\textnormal{PearsonGAN}}$ = $\mathbb{E}[D(\mathbf{x})] + \mathbb{E}[\frac{1}{4}D(G(\mathbf{z}))^{2} + D(G(\mathbf{z}))]$ 
    & 
    $\mathcal{L}^{\textnormal{TVGAN}}$ = $-\frac{1}{2}\mathbb{E}[\tanh (D(\mathbf{x}))] + \frac{1}{2}\mathbb{E}[\tanh (D(G(\mathbf{z})))]$  \\ \midrule
    
    $\mathcal{L}^{\textnormal{ForwGAN}}$ = $\mathbb{E}[D(\mathbf{x})] + \mathbb{E}[\textnormal{exp}(D(G(\mathbf{z})))-1]$ 
    & 
    $\mathcal{L}^{\textnormal{RevGAN}}$ = $\mathbb{E}[-\textnormal{exp}(D(\mathbf{x}))] + \mathbb{E}[-1 - (D(G(\mathbf{z})))]$  \\ \midrule
    
    $\mathcal{L}^{\textnormal{HellingerGAN}}$ = $\mathbb{E}[1 - \textnormal{exp}(-D(\mathbf{x}))] + \mathbb{E}[\frac{1 - \textnormal{exp}(D(G(\mathbf{z})))}{\textnormal{exp}(D(G(\mathbf{z})))}]$ & $\mathcal{L}^{\textnormal{JSGAN}}$ = $\mathbb{E}[2 - (1+\textnormal{exp}(-D(\mathbf{x})))] - \mathbb{E}[2 - \textnormal{exp}(D(G(\mathbf{z})))]$  \\ \bottomrule

  \end{tabular}}
   \caption{\label{tab:losses} The loss function for each GAN variant with slight abuse of parameterization notation on the expectations, $G$, and $D$. Note that $G$ is parameterized by $\theta_{G}$, $D$ is parameterized by $\theta_{D}$, $x \sim p_{d}$, $z \sim p_{g}$, $\delta \sim \mathcal{N}(0,cI)$, $\Hat{\Omega}(D, G) =  \frac{1}{2}\mathbb{E}[D(\mathbf{x})]^{2} - \frac{1}{2}\mathbb{E}[D(G(\mathbf{z}))]^{2}$, $D_{AE}$ indicates that $D$ is an autoencoder, $\mathbf{c}=[\mathbf{c_1}, \mathbf{c_2}]$ are structured latent variables where $\mathbf{c}'$ is sampled from the approximated distribution $p_{\mathbf{c}}(\mathbf{c}|\mathbf{x})$, $\nabla_{(\mathbf{\cdot})}$ is the gradient of the loss with respect to $(\mathbf{\cdot})$, and $k_{t}$, $\lambda$, and $\rho$ are introduced hyperparameters.} \label{tab:gansfncs}
\end{table*}

In our implementations, both $D$ and $G$ consist of two feedforward network layers each; the full architecture has four layers total. We apply a ReLU activation function to the output of each layer and sample the noise prior $\mathbf{z}$ from $\mathcal{N}(0,\frac{h}{4}I)$, where $h$ is the hidden dimension size. All models have the same number of trainable parameters except InfoGAN and BEGAN due to their use of latent variables as inputs to $D$ and formation of $D$ as an autoencoder, respectively. \citet{infogan} argue that this difference is negligible for InfoGAN and we do not observe that it gives BEGAN any tangible advantage over other models. Trainable parameter counts can be found in Table \ref{tab:modelparams}.

\subsection*{Data}

We train each of these variants by randomly sampling 1,000, 10,000, and 100,000 data points from the following six explicitly parameterized multivariate distributions: Gaussian with mean $\mathbf{\mu}$ and symmetric, full rank covariance $\mathbf{\Sigma}$ both from $[0,1]$; exponential with inverse mean shape $\mathbf{\lambda}$ from $[0,1]$; beta with shape parameters $a$ and $\beta$ both from $[0,1]$; gamma with shape $k$ from $[0,10]$ and scale $\theta$ from $[0,2]$; Gumbel with location $\mu$ and scale $\beta$ both from $[0,1]$; and Laplace with location $\mu$ and scale $\beta$ both from $[0,1]$. For each of these distributions and numbers of samples, we generate datasets of 16, 32, 64, and 128 dimensions. We note that by the Universal Approximation Theorem, our proposed network architecture should be able to model each of these distributions without exception.


\subsection*{Hyperparameters}
For all models and data distributions, we conduct 20 grid search trials with random network initializations for learning rates $\gamma \in [2\text{e}^{-1}, 2\text{e}^{-2}, 2\text{e}^{-3}]$, hidden dimension sizes $h \in [32, 64, 128, 256, 512]$, and batch size $b = 1024$.\footnote{We also ran full experiments for $b \in [128, 256, 512]$, but limit our analyses to $b$ = 1024 as results across different batch sizes are not comparable due to greater noise in the data generation process at lower values of $b$.} For models with introduced hyperparameters, we use those given in the original the paper. We use the Adam optimizer with default settings \cite{adam} and train for 25 epochs.

\subsection*{Measuring Divergence}
We evaluate the difference between $p_{d}$ and $p_g$ using Kullback-Leibler divergence (KL), Jensen-Shannon divergence (JS), and Wasserstein Distance (WD). KL and JS focus on the alignment of the modes of the distributions and WD emphasizes how much $p_g$ must be modified to reach $p_{d}$. Whereas JS and WD are symmetric, KL is not. For any of these measures, a value of 0 can be interpreted as indicating the two distributions being compared are identical \cite{jensenshannon, kullbackleibler,villani2008optimal}. We report results as the divergence between a generated batch and a test batch of size $b=1024$ at the end of every epoch.

\subsection*{Estimating $p_g$}
Although we have access to the true data distribution $p_{d}$, we must estimate the probability distribution of $p_g$. Since the data dimensionality is low, we construct a dimension-wise histogram for each data point.\footnote{Kernel density estimation was found to give similar outputs while being more computationally expensive.} In doing so, we assume that each dimension is independent from the others. This assumption is valid in the case of all experiments involving non-Gaussian data, which follows from the multivariate model being a product of the marginal distributions. To select the optimal bin width $B_w$ in the histogram, we follow the Freedman-Diaconis rule: $B_w = \frac{2 \cdot IQR({\bf \Tilde{x}})}{\sqrt[3]{M}}$, where $IQR$ is the inter-quartile range of the $M$ samples ${\bf \Tilde{x}} = \{x_1, ..., x_M\}$ from the distribution being approximated.
This initialization minimizes the difference between the areas under the empirical and theoretical probability distributions~\cite{diaconis}. 
\section{Results}

\begin{figure*}[!htb]
\centering
    \includegraphics[width=0.96\linewidth]{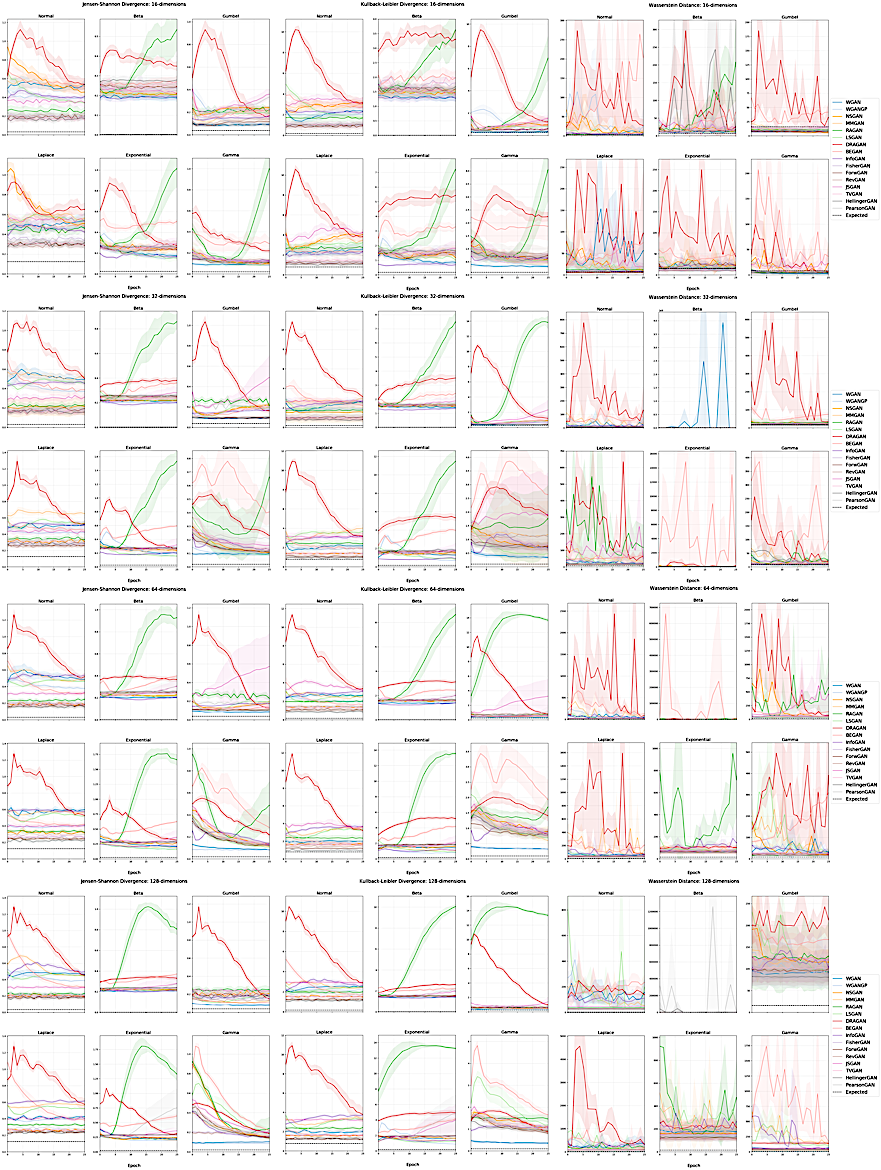}

    \caption{Performance of GAN variants for their best hyperparameter settings, respectively, trained on 100,000 samples across normal, beta, Gumbel, Laplace, exponential, and gamma multivariate distributions of dimensionalities $N=16$ (rows 1 and 2), $N=32$ (rows 3 and 4), $N=64$ (rows 5 and 6), and $N=128$ (rows 7 and 8). Plots display metric performance as a function of epoch. Shaded areas represent the region of the 95\% confidence interval of the respective model computed over 20 trials. ``Expected'' indicates the empirical average divergence of a generated batch of size $b=1024$ for the given distribution across 20 trials where samples are generated independently, i.e. one at a time. Best viewed in color.}
    \label{fig:lotsofgraphs100000}
\end{figure*}
\begin{figure*}[!htb]
\centering
    \includegraphics[width=0.99\linewidth]{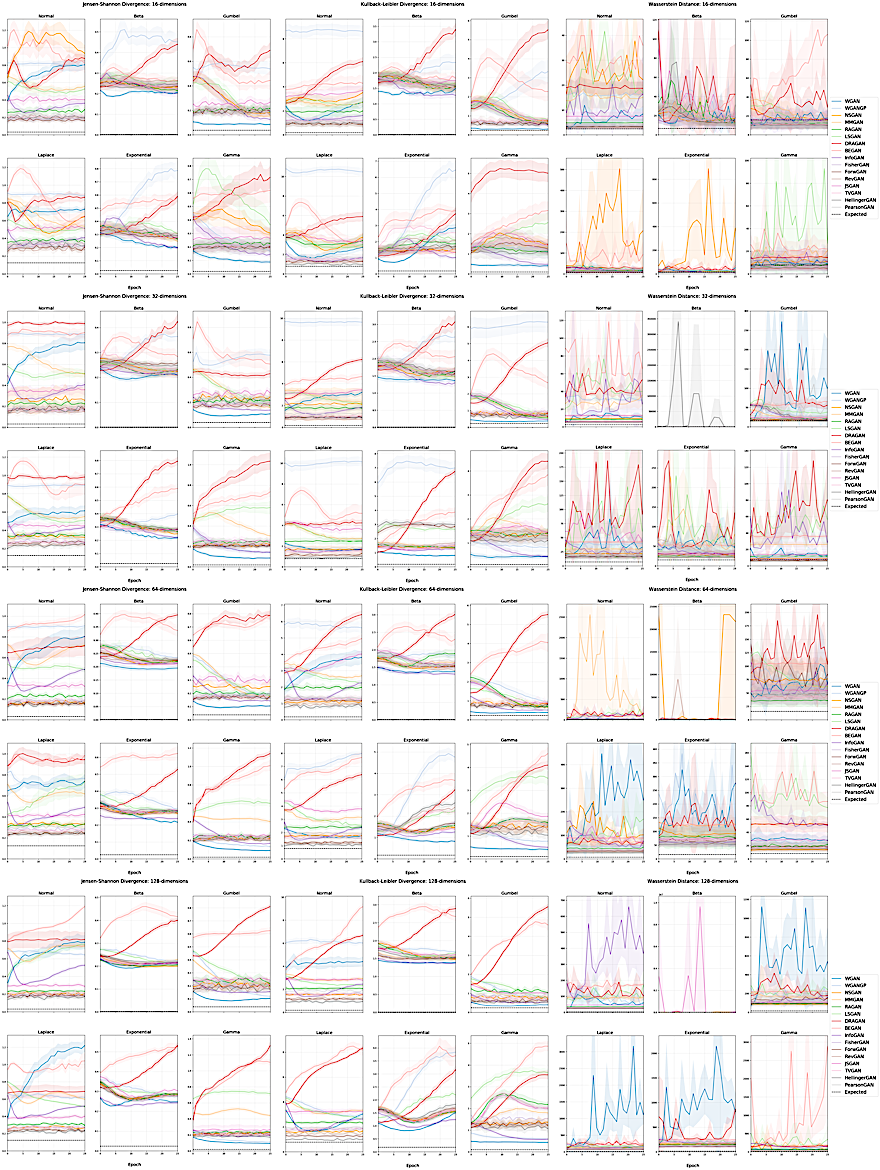}

    \caption{Performance of GAN variants for their best hyperparameter settings, respectively, trained on 10,000 samples across normal, beta, Gumbel, Laplace, exponential, and gamma multivariate distributions of dimensionalities $N=16$ (rows 1 and 2), $N=32$ (rows 3 and 4), $N=64$ (rows 5 and 6), and $N=128$ (rows 7 and 8). Plots display metric performance as a function of epoch. Shaded areas represent the region of the 95\% confidence interval of the respective model computed over 20 trials. ``Expected'' indicates the empirical average divergence of a generated batch of size $b=1024$ for the given distribution across 20 trials where samples are generated independently, i.e. one at a time. Best viewed in color.}
    \label{fig:lotsofgraphs10000}
\end{figure*}
\begin{figure*}[!htb]
\centering
    \includegraphics[width=0.99\linewidth]{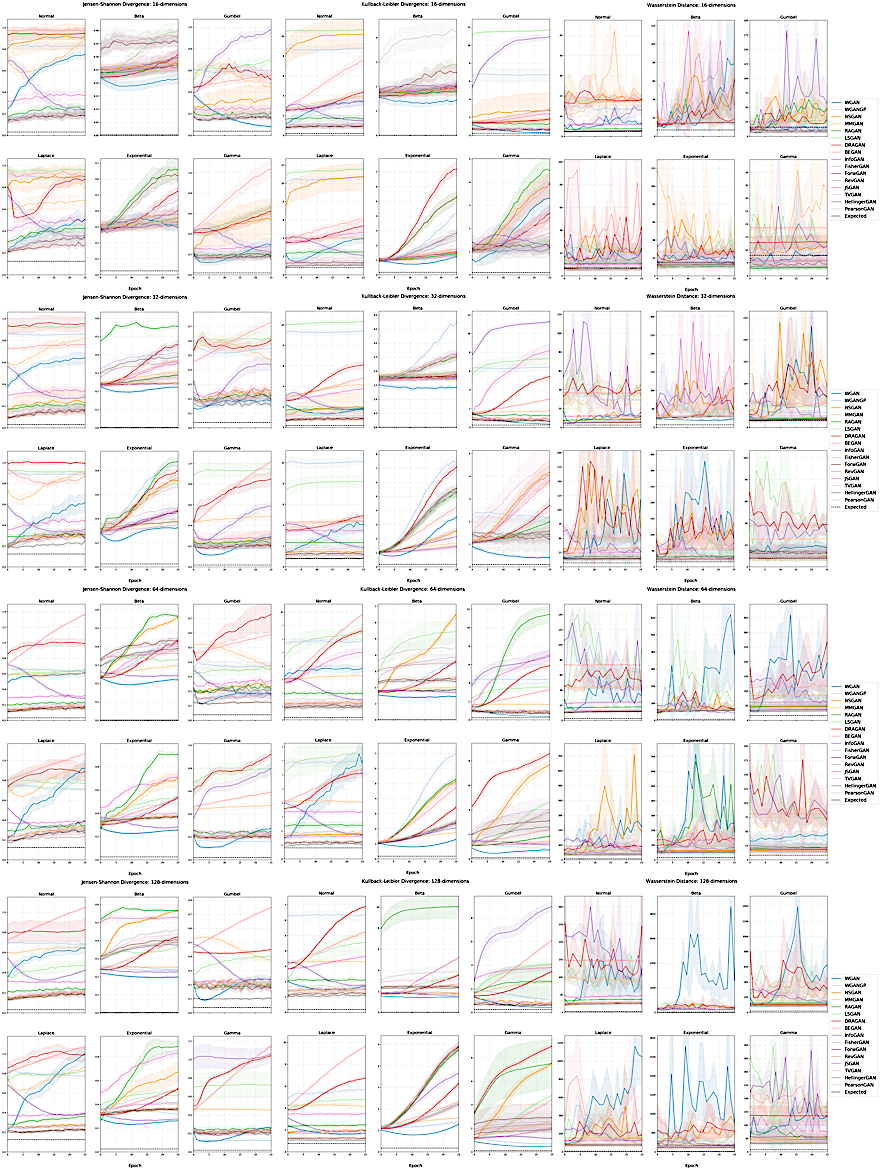}
    
    \caption{Performance of GAN variants for their best hyperparameter settings, respectively, trained on 1,000 samples across normal, beta, Gumbel, Laplace, exponential, and gamma multivariate distributions of dimensionalities $N=16$ (rows 1 and 2), $N=32$ (rows 3 and 4), $N=64$ (rows 5 and 6), and $N=128$ (rows 7 and 8). Plots display metric performance as a function of epoch. Shaded areas represent the region of the 95\% confidence interval of the respective model computed over 20 trials. ``Expected'' indicates the empirical average divergence of a generated batch of size $b=1024$ for the given distribution across 20 trials where samples are generated independently, i.e. one at a time. Best viewed in color.}
    \label{fig:lotsofgraphs1000}
\end{figure*}
\begin{figure*}[!htb]
    \centering
    \includegraphics[height=0.83\textheight,width=0.99\textwidth,keepaspectratio]{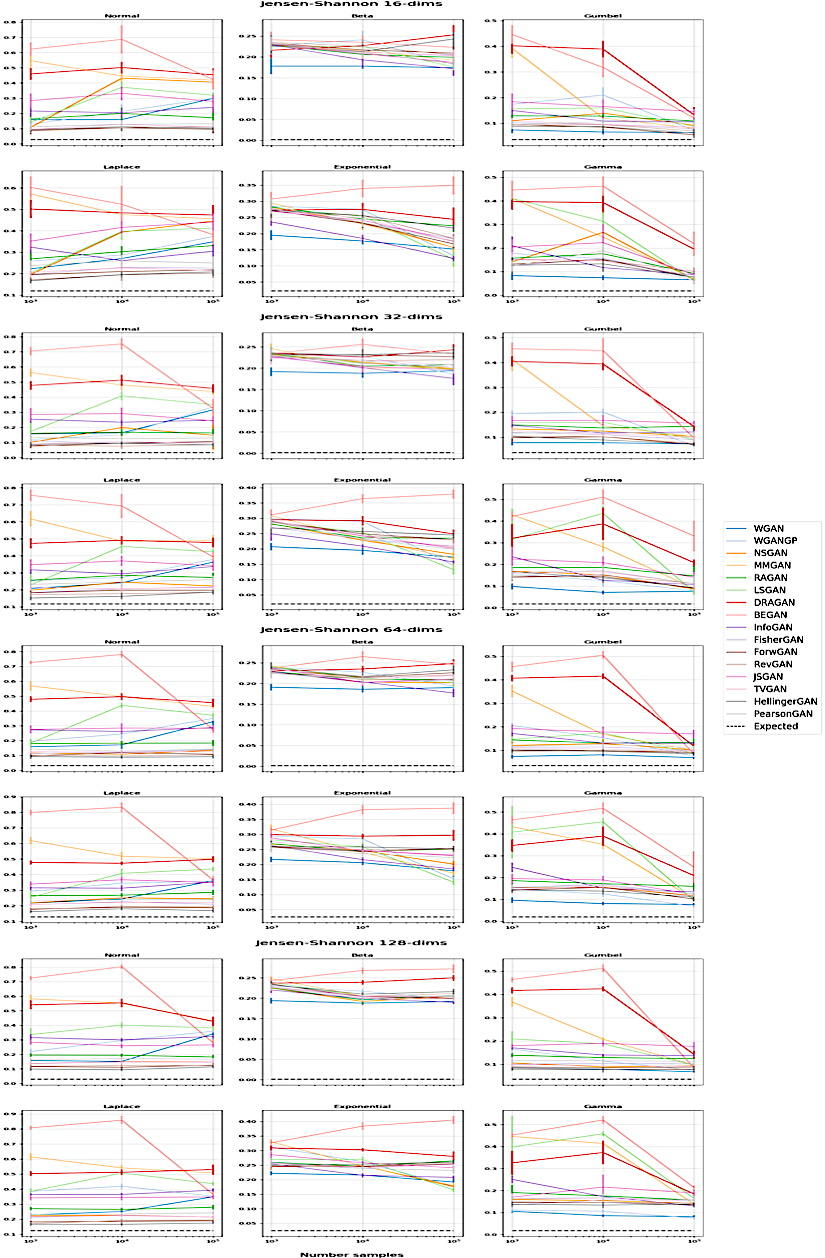}
     
    \caption{Confidence intervals of the Jensen-Shannon divergence performances for the best performing hyperparameter over 20 trials as a function of sample size for dimensionalities $N=16$ (rows 1 and 2), $N=32$ (rows 3 and 4), $N=64$ (rows 5 and 6), and $N=128$ (rows 7 and 8). ``Expected'' indicates the empirical average divergence of a generated batch of size $b=1024$ for the given distribution would be across 20 trials where samples are generated independently, i.e. one at a time. Best viewed in color.}
    \label{fig:fncsamples_JS}
\end{figure*}
\begin{figure*}[!htb]
    \centering
    \includegraphics[height=0.83\textheight,width=0.99\textwidth,keepaspectratio]{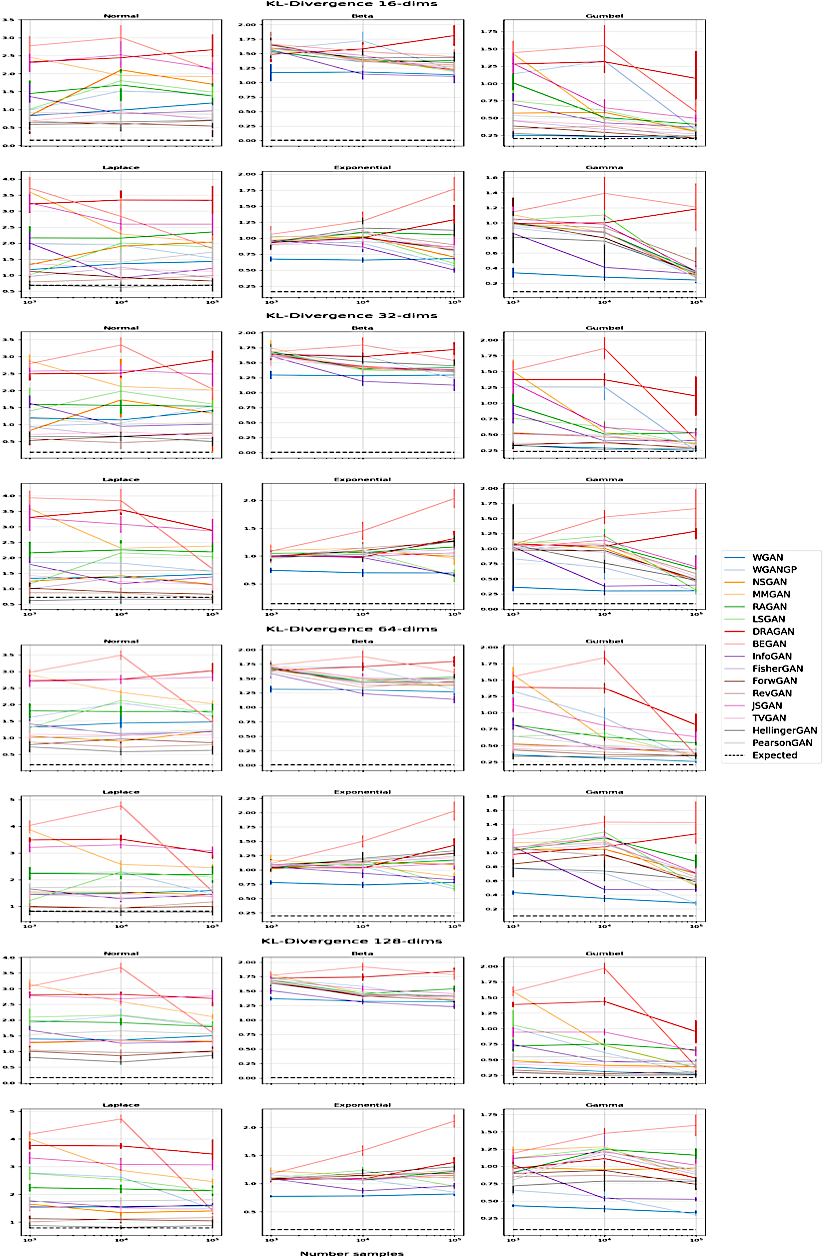}
     
    \caption{Confidence intervals of the Kullback-Leibler divergence performances for the best performing hyperparameter over 20 trials as a function of sample size for dimensionalities $N=16$ (rows 1 and 2), $N=32$ (rows 3 and 4), $N=64$ (rows 5 and 6), and $N=128$ (rows 7 and 8). ``Expected'' indicates the empirical average divergence of a generated batch of size $b=1024$ for the given distribution would be across 20 trials where samples are generated independently, i.e. one at a time. Best viewed in color.}
    \label{fig:fncsamples_KL}
\end{figure*}
\begin{figure*}[!htb]
    \centering
    \includegraphics[height=0.83\textheight,width=0.99\textwidth,keepaspectratio]{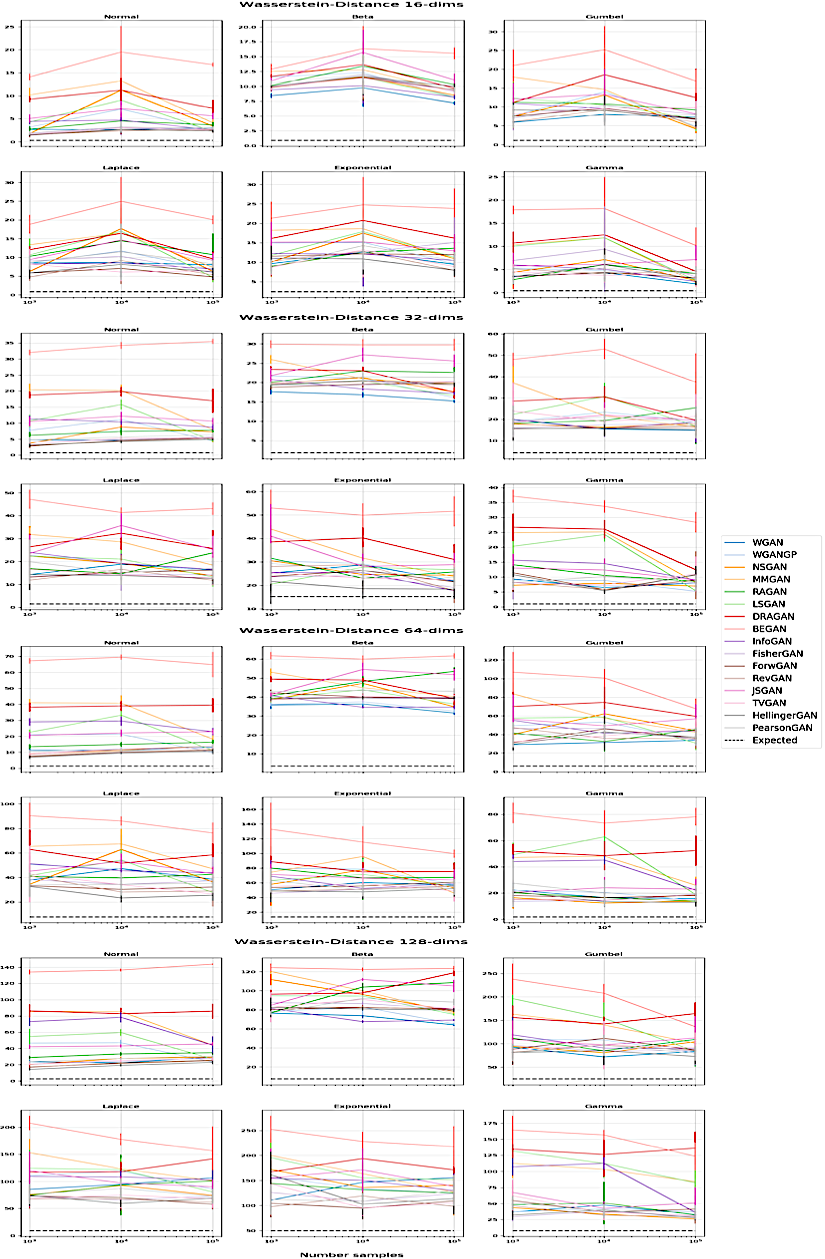}
     
    \caption{Confidence intervals of the Wasserstein Distance performances for the best performing hyperparameter over 20 trials as a function of sample size for dimensionalities $N=16$ (rows 1 and 2), $N=32$ (rows 3 and 4), $N=64$ (rows 5 and 6), and $N=128$ (rows 7 and 8). ``Expected'' indicates the empirical average divergence of a generated batch of size $b=1024$ for the given distribution would be across 20 trials where samples are generated independently, i.e. one at a time. Best viewed in color.}
    \label{fig:fncsamples_WD}
\end{figure*}

In our analyses, we take the same approach as \citet{lucic2017gans} and \citet{divergencetraining}: we let the ``best'' hyperparameter setting be the one that achieved the lowest minimum performance on average across all trials for each distribution, metric, and number of training samples, respectively. We include results, visualizations, and evidence to support all conclusions in the appendices. For the best hyperparameter settings in our learning environment, we find that:
\begin{enumerate}

\item \textbf{GANs exhibit similar learning trends across dimensionalities}: For many of the models, performance under the best hyperparameter setting consistently follows a trend across dimensionalities for all three tested metrics. At the same time, performance generally worsens with increased dimensionality. See Figures \ref{fig:lotsofgraphs100000}, \ref{fig:lotsofgraphs10000}, \ref{fig:lotsofgraphs1000}.

\item \textbf{Learning depends on the underlying distribution and its complexity:} Models which do well on some distributions perform poorly on the same distribution with higher dimensionality, or on other distributions of the same dimensionality. It is not immediately apparent that these differences are due to model design. \citet{lucic2017gans} make a similar finding in the case of image datasets with varying complexities. See Figures \ref{fig:lotsofgraphs100000}, \ref{fig:lotsofgraphs10000}, \ref{fig:lotsofgraphs1000}.

\item \textbf{Number of training samples can have a large impact on performance:} Some GAN variants are able to achieve the same performance learning from 1,000 samples as 10,000 or 100,000 samples, while others show large performance jumps with increased amounts of data. At the same time, almost all GAN variants begin to worsen in performance within five epochs for 1,000 training samples. The number of training samples seems to be critical to some models' performances, which was also noted by \citet{divergencetraining}. See Figures \ref{fig:fncsamples_JS}, \ref{fig:fncsamples_KL}, and \ref{fig:fncsamples_WD}.

\item \textbf{Evaluation and comparison are metric-dependent:} Relative ranking of GAN variants according to performance varies depending on the evaluation metric used to rank them. No single GAN performed best across all metrics for any dataset or dimensionality. We concur with previous studies that GANs generally perform the same, although there are variants that perform worse than others on some distributions \cite{borji2018pros,huang2018an,divergencetraining,lucic2017gans,santurkar}. We warn against ranking models as relative differences can be marginal. See Tables \ref{tab:metricdependent}, \ref{tab:jsconfidenceintervals}, \ref{tab:klconfidenceintervals}, and \ref{tab:wdconfidenceintervals}.

\item \textbf{Diverse sets of hyperparameters can produce a ``best'' result:} Many, diverse hyperparameter settings yielded superior performances to the best average minimum performance, but these models did not achieve those minima with tight confidence bounds. Furthermore, we see that even on the best performing hyperparameter settings, our tested models preferred widely different hidden dimensionalities and learning rates; some variants with less parameters outperformed others that had more. We agree with previous work that it is important to present results that are able to be consistently reproduced \cite{borji2018pros, divergencetraining, lucic2017gans, lee2017ability, theis}. See Table \ref{tab:klminhyperparams}.

\item \textbf{Some GANs are more robust to hyperparameter changes than others:} With respect to the distribution, dimensionality, and training set size being approximated, some models yielded average minimum performances for more hyperparameters than others that fell within the confidence interval of the best average minimum performance under consideration. This is an indication that some GANs can perform well under a greater range of hyperparameter settings than others. See Table \ref{tab:robust}.
\end{enumerate}

\section{Future Work}

In future work, we plan to analyze cases where GAN variants underperform relative to others and relate the characteristics of the distribution being modeled to the assumptions made in designing the variant, e.g. by empirically considering whether a normally distributed prior hurts performance on non-normal distributions. We would also like to use longer training times and more complex models to evaluate additional synthetic datasets such as multivariate mixture models, colored circles, and autoencoded image datasets.

\section*{Acknowledgements}
We would like to thank Deb Roy, Iyad Rahwan, Manolis Zampetakis, and the Media Lab Consortium for their support of this research. Aleksander M\k{a}dry and Costis Daskalakis provided us with inspiration to pursue this project. Shayne O'Brien is partially funded through a National Science Foundation Graduate Research Fellowship.

\bibliographystyle{plainnat}

\newpage
\appendix 

\section{GANs exhibit similar learning trends across dimensionalities, and learning depends on the underlying distribution and its complexity}

\clearpage

\section{Number of training samples can have a large impact on performance}

\clearpage

\section{Evaluation and comparisons are metric-dependent} 
\begin{table*}[!htb]
  \centering
  \scriptsize
    \begin{tabular}{lllllll}
     Model        & Normal       & Beta         & Gumbel       & Laplace      & Exponential   & Gamma               \\ \toprule
     
     WGAN                & [7, 7, 13]   & [0, 0, 5]    & [0, 0, 1]    & [5, 7, 13]   & [0, 0, 3]     & [0, 0, 2]    \\
                         & [8, 9, 10]   & [0, 1, 4]    & [9, 5, 7]    & [9, 9, 9]    & [1, 3, 4]     & [0, 0, 1]    \\
                         & [7, 2, 11]   & [0, 0, 4]    & [1, 2, 9]    & [8, 8, 11]   & [3, 6, 4]     & [5, 0, 2]    \\ \midrule
                         
     WGANGP              & [8, 10, 14]  & [13, 2, 0]   & [13, 8, 4]   & [9, 9, 14]   & [13, 1, 0]    & [9, 1, 0]    \\
                         & [9, 10, 11]  & [11, 3, 1]   & [13, 12, 9]  & [12, 11, 10] & [12, 4, 1]    & [9, 4, 0]    \\
                         & [9, 9, 0]    & [2, 1, 1]    & [8, 11, 3]   & [10, 10, 2]  & [9, 4, 3]     & [10, 3, 0]   \\ \midrule

     NSGAN               & [4, 6, 6]    & [6, 3, 4]    & [2, 4, 6]    & [2, 8, 5]    & [5, 4, 6]     & [6, 2, 3]    \\
                         & [5, 1, 5]    & [7, 2, 3]    & [1, 4, 4]    & [5, 4, 5]    & [2, 0, 5]     & [7, 8, 5]    \\
                         & [4, 7, 10]   & [10, 4, 6]   & [5, 6, 1]    & [4, 9, 3]    & [4, 5, 10]    & [6, 4, 4]    \\ \midrule
                         
     MMGAN               & [14, 14, 11] & [7, 5, 6]    & [12, 11, 7]  & [14, 13, 11] & [3, 5, 4]     & [13, 5, 4]   \\
                         & [14, 13, 12] & [10, 6, 5]   & [12, 9, 10]  & [13, 13, 13] & [14, 11, 6]   & [14, 9, 2]   \\
                         & [14, 13, 12] & [14, 5, 5]   & [14, 4, 2]   & [14, 11, 5]  & [14, 13, 12]  & [13, 12, 1]  \\ \midrule
                  
     RAGAN               & [10, 8, 7]   & [11, 13, 14] & [8, 9, 13]   & [10, 10, 9]  & [6, 13, 14]   & [10, 9, 14]  \\
                         & [6, 7, 7]    & [12, 13, 14] & [7, 8, 8]    & [6, 8, 7]    & [11, 14, 14]  & [10, 6, 11]  \\ 
                         & [8, 10, 13]  & [8, 15, 13]  & [4, 12, 10]  & [7, 4, 13]   & [2, 8, 13]    & [0, 6, 12]   \\ \midrule
                  
     LSGAN               & [9, 12, 8]   & [9, 4, 2]    & [9, 10, 8]   & [6, 12, 8]   & [2, 2, 1]     & [14, 7, 1]   \\ 
                         & [10, 12, 9]  & [9, 4, 0]    & [11, 11, 11] & [8, 12, 11]  & [13, 10, 3]   & [12, 11, 3]  \\
                         & [11, 8, 1]   & [11, 3, 3]   & [11, 8, 0]   & [11, 14, 0]  & [11, 12, 7]   & [12, 13, 5]  \\ \midrule
     
     DRAGAN              & [13, 13, 10] & [14, 14, 1]  & [14, 15, 11] & [13, 15, 10] & [14, 14, 2]   & [12, 14, 11] \\ 
                         & [13, 14, 13] & [14, 14, 10] & [14, 15, 15] & [14, 15, 15] & [10, 13, 10]  & [11, 14, 14] \\
                         & [13, 14, 7]  & [5, 7, 0]    & [13, 14, 4]  & [13, 13, 10] & [13, 14, 0]   & [14, 14, 6]  \\ \midrule
     
     BEGAN               & [15, 15, 15] & [15, 15, 15] & [15, 14, 15] & [15, 14, 15] & [15, 15, 15]  & [15, 15, 15] \\
                         & [15, 15, 15] & [15, 15, 15] & [15, 14, 14] & [15, 14, 12] & [15, 15, 15]  & [15, 15, 15] \\
                         & [15, 15, 15] & [15, 14, 14] & [15, 15, 15] & [15, 15, 15] & [15, 15, 15]  & [15, 15, 15] \\ \midrule
     
     InfoGAN             & [11, 9, 9]   & [1, 1, 3]    & [10, 12, 14] & [11, 6, 6]   & [1, 3, 5]     & [4, 3, 5]    \\ 
                         & [11, 8, 8]   & [1, 0, 2]    & [10, 10, 13] & [10, 5, 8]   & [6, 1, 0]     & [8, 1, 7]    \\
                         & [10, 11, 4]  & [1, 2, 2]    & [9, 7, 7]    & [9, 5, 8]    & [7, 7, 8]     & [11, 10, 3]  \\ \midrule
     
     FisherGAN           & [5, 2, 5]    & [4, 8, 8]    & [5, 5, 5]    & [3, 4, 4]    & [10, 8, 11]   & [5, 10, 7]   \\ 
                         & [2, 4, 4]    & [5, 10, 7]   & [2, 6, 5]    & [2, 6, 1]    & [4, 2, 8]     & [3, 5, 10]   \\
                         & [5, 5, 9]    & [7, 12, 7]   & [0, 0, 11]   & [6, 7, 6]    & [8, 9, 11]    & [8, 5, 9]    \\ \midrule
     
     ForwGAN             & [0, 3, 0]    & [5, 9, 10]   & [3, 1, 0]    & [4, 3, 2]    & [12, 6, 7]    & [1, 6, 8]    \\ 
                         & [0, 0, 6]    & [3, 8, 11]   & [4, 7, 1]    & [3, 1, 3]    & [8, 9, 9]     & [4, 7, 8]    \\
                         & [0, 1, 3]    & [3, 8, 10]   & [7, 10, 8]   & [2, 2, 1]    & [10, 3, 5]    & [7, 8, 8]    \\ \midrule
    
     RevGAN              & [2, 4, 2]    & [2, 10, 9]   & [4, 6, 2]    & [1, 0, 1]    & [4, 9, 12]    & [2, 4, 12]   \\ 
                         & [4, 3, 1]    & [4, 7, 6]    & [5, 3, 2]    & [1, 2, 2]    & [7, 5, 13]    & [2, 2, 13]   \\
                         & [2, 0, 8]    & [6, 9, 11]   & [2, 9, 5]    & [0, 1, 4]    & [5, 1, 1]     & [2, 1, 13]   \\ \midrule
     
     JSGAN               & [12, 11, 12] & [12, 11, 11] & [11, 13, 12] & [12, 11, 12] & [11, 10, 8]   & [11, 13, 13] \\ 
                         & [12, 11, 14] & [13, 12, 13] & [8, 13, 12]  & [11, 10, 14] & [9, 12, 11]   & [13, 12, 12] \\
                         & [12, 12, 14] & [13, 10, 15] & [12, 13, 13] & [12, 12, 14] & [12, 11, 14]  & [9, 11, 14]  \\ \midrule
     
     TVGAN               & [3, 0, 3]    & [8, 6, 12]   & [6, 2, 9]    & [7, 2, 3]    & [7, 7, 9]     & [7, 11, 6]   \\ 
                         & [3, 5, 3]    & [6, 9, 9]    & [6, 2, 6]    & [7, 7, 4]    & [3, 7, 7]     & [6, 10, 4]   \\
                         & [3, 4, 5]    & [9, 6, 8]    & [10, 5, 14]  & [5, 0, 9]    & [0, 0, 6]     & [1, 2, 11]   \\ \midrule
     
     HellingerGAN        & [1, 1, 1]    & [3, 7, 7]    & [1, 3, 3]    & [0, 1, 0]    & [8, 12, 13]   & [3, 8, 9]    \\ 
                         & [1, 2, 0]    & [2, 5, 8]    & [0, 1, 0]    & [0, 0, 0]    & [5, 8, 12]    & [1, 3, 9]    \\
                         & [1, 3, 2]    & [4, 11, 12]  & [3, 3, 6]    & [3, 3, 7]    & [6, 2, 2]     & [4, 7, 7]    \\ \midrule
     
     PearsonGAN          & [6, 5, 4]    & [10, 12, 13] & [7, 7, 10]   & [8, 5, 7]    & [9, 11, 10]   & [8, 12, 10]  \\ 
                         & [7, 6, 2]    & [8, 11, 12]  & [3, 0, 3]    & [4, 3, 6]    & [0, 6, 2]     & [5, 13, 6]   \\
                         & [6, 6, 6]    & [12, 13, 9]  & [6, 1, 12]   & [1, 6, 12]   & [1, 10, 9]    & [3, 9, 10]   \\ \bottomrule
    \end{tabular}
  \caption{Performance-based relative ranking of GAN variants for 1,000 samples (first entry of each row), 10,000 samples (second entry), and 100,000 samples (third entry) with respect to Jensen-Shannon divergence (first row), Kullback-Leibler divergence (second row), and Wasserstein distance (third row) for the best hyperparameter settings for dimensionality $N=128$. Note that differences between performances are sometimes marginal. Dimensionalities $N \in \left\{16, 32, 64 \right\}$ yielded similar ranking results to these with slight variations.} \label{tab:metricdependent}
\end{table*}

\begin{table*}[!htb]
  \centering
  \resizebox{\textwidth}{!}{%
    \begin{tabular}{lllllll}
     Model        & Normal       & Beta         & Gumbel       & Laplace      & Exponential   & Gamma               \\ \toprule
    WGAN                & 0.128 $\pm$ 0.019 & 0.099 $\pm$ 0.010 & 0.062 $\pm$ 0.007 & 0.201 $\pm$ 0.019 & 0.156 $\pm$ 0.010 & 0.078 $\pm$ 0.011 \\
                        & 0.160 $\pm$ 0.029 & 0.109 $\pm$ 0.011 & 0.066 $\pm$ 0.009 & 0.271 $\pm$ 0.027 & 0.128 $\pm$ 0.017 & 0.059 $\pm$ 0.009 \\ 
                        & 0.299 $\pm$ 0.037 & 0.114 $\pm$ 0.018 & 0.064 $\pm$ 0.014 & 0.349 $\pm$ 0.050 & 0.108 $\pm$ 0.013 & 0.062 $\pm$ 0.013 \\ \midrule
    WGANGP              & 0.129 $\pm$ 0.026 & 0.190 $\pm$ 0.020 & 0.174 $\pm$ 0.023 & 0.259 $\pm$ 0.042 & 0.248 $\pm$ 0.019 & 0.143 $\pm$ 0.032 \\
                        & 0.213 $\pm$ 0.048 & 0.121 $\pm$ 0.011 & 0.092 $\pm$ 0.013 & 0.287 $\pm$ 0.033 & 0.122 $\pm$ 0.014 & 0.068 $\pm$ 0.012 \\
                        & 0.291 $\pm$ 0.022 & 0.066 $\pm$ 0.016 & 0.073 $\pm$ 0.011 & 0.341 $\pm$ 0.012 & 0.083 $\pm$ 0.010 & 0.051 $\pm$ 0.011 \\ \midrule
    NSGAN               & 0.106 $\pm$ 0.021 & 0.161 $\pm$ 0.019 & 0.079 $\pm$ 0.009 & 0.197 $\pm$ 0.022 & 0.213 $\pm$ 0.013 & 0.125 $\pm$ 0.025 \\
                        & 0.120 $\pm$ 0.018 & 0.120 $\pm$ 0.013 & 0.085 $\pm$ 0.012 & 0.277 $\pm$ 0.028 & 0.146 $\pm$ 0.015 & 0.089 $\pm$ 0.020 \\
                        & 0.126 $\pm$ 0.025 & 0.105 $\pm$ 0.013 & 0.080 $\pm$ 0.013 & 0.242 $\pm$ 0.021 & 0.110 $\pm$ 0.014 & 0.060 $\pm$ 0.016 \\ \midrule
    MMGAN               & 0.422 $\pm$ 0.040 & 0.168 $\pm$ 0.014 & 0.148 $\pm$ 0.026 & 0.444 $\pm$ 0.033 & 0.202 $\pm$ 0.013 & 0.311 $\pm$ 0.037 \\
                        & 0.401 $\pm$ 0.025 & 0.142 $\pm$ 0.032 & 0.100 $\pm$ 0.009 & 0.440 $\pm$ 0.036 & 0.142 $\pm$ 0.018 & 0.098 $\pm$ 0.023 \\
                        & 0.246 $\pm$ 0.033 & 0.110 $\pm$ 0.027 & 0.081 $\pm$ 0.009 & 0.334 $\pm$ 0.042 & 0.107 $\pm$ 0.015 & 0.060 $\pm$ 0.009 \\ \midrule
    RaGAN               & 0.160 $\pm$ 0.029 & 0.180 $\pm$ 0.019 & 0.100 $\pm$ 0.013 & 0.269 $\pm$ 0.027 & 0.216 $\pm$ 0.023 & 0.146 $\pm$ 0.035 \\
                        & 0.169 $\pm$ 0.027 & 0.184 $\pm$ 0.016 & 0.099 $\pm$ 0.021 & 0.284 $\pm$ 0.023 & 0.220 $\pm$ 0.015 & 0.128 $\pm$ 0.028 \\
                        & 0.156 $\pm$ 0.029 & 0.199 $\pm$ 0.016 & 0.104 $\pm$ 0.014 & 0.293 $\pm$ 0.029 & 0.224 $\pm$ 0.019 & 0.092 $\pm$ 0.020 \\ \midrule
    LSGAN               & 0.142 $\pm$ 0.035 & 0.170 $\pm$ 0.025 & 0.115 $\pm$ 0.015 & 0.204 $\pm$ 0.017 & 0.198 $\pm$ 0.019 & 0.340 $\pm$ 0.052 \\
                        & 0.306 $\pm$ 0.021 & 0.127 $\pm$ 0.014 & 0.094 $\pm$ 0.019 & 0.333 $\pm$ 0.022 & 0.128 $\pm$ 0.020 & 0.102 $\pm$ 0.027 \\
                        & 0.206 $\pm$ 0.016 & 0.080 $\pm$ 0.026 & 0.088 $\pm$ 0.010 & 0.288 $\pm$ 0.019 & 0.090 $\pm$ 0.012 & 0.056 $\pm$ 0.014 \\ \midrule
    DRAGAN              & 0.354 $\pm$ 0.041 & 0.212 $\pm$ 0.025 & 0.272 $\pm$ 0.031 & 0.392 $\pm$ 0.032 & 0.260 $\pm$ 0.021 & 0.301 $\pm$ 0.045 \\ 
                        & 0.385 $\pm$ 0.064 & 0.198 $\pm$ 0.026 & 0.248 $\pm$ 0.051 & 0.457 $\pm$ 0.037 & 0.259 $\pm$ 0.054 & 0.216 $\pm$ 0.035 \\
                        & 0.239 $\pm$ 0.017 & 0.069 $\pm$ 0.020 & 0.098 $\pm$ 0.010 & 0.297 $\pm$ 0.021 & 0.096 $\pm$ 0.020 & 0.087 $\pm$ 0.016 \\ \midrule
    BEGAN               & 0.549 $\pm$ 0.050 & 0.231 $\pm$ 0.018 & 0.329 $\pm$ 0.069 & 0.507 $\pm$ 0.051 & 0.301 $\pm$ 0.026 & 0.446 $\pm$ 0.040 \\ 
                        & 0.457 $\pm$ 0.056 & 0.226 $\pm$ 0.016 & 0.179 $\pm$ 0.057 & 0.445 $\pm$ 0.050 & 0.340 $\pm$ 0.027 & 0.287 $\pm$ 0.061 \\
                        & 0.348 $\pm$ 0.055 & 0.223 $\pm$ 0.017 & 0.120 $\pm$ 0.025 & 0.381 $\pm$ 0.041 & 0.307 $\pm$ 0.025 & 0.218 $\pm$ 0.052 \\ \midrule
    InfoGAN             & 0.190 $\pm$ 0.023 & 0.122 $\pm$ 0.019 & 0.113 $\pm$ 0.012 & 0.261 $\pm$ 0.024 & 0.151 $\pm$ 0.011 & 0.119 $\pm$ 0.019 \\
                        & 0.204 $\pm$ 0.028 & 0.105 $\pm$ 0.011 & 0.104 $\pm$ 0.007 & 0.261 $\pm$ 0.026 & 0.134 $\pm$ 0.014 & 0.089 $\pm$ 0.019 \\
                        & 0.202 $\pm$ 0.032 & 0.085 $\pm$ 0.015 & 0.104 $\pm$ 0.005 & 0.243 $\pm$ 0.022 & 0.102 $\pm$ 0.010 & 0.066 $\pm$ 0.018 \\ \midrule
    FisherGAN           & 0.112 $\pm$ 0.030 & 0.156 $\pm$ 0.013 & 0.083 $\pm$ 0.013 & 0.197 $\pm$ 0.029 & 0.223 $\pm$ 0.020 & 0.117 $\pm$ 0.023 \\
                        & 0.109 $\pm$ 0.027 & 0.160 $\pm$ 0.011 & 0.081 $\pm$ 0.010 & 0.221 $\pm$ 0.026 & 0.193 $\pm$ 0.013 & 0.124 $\pm$ 0.024 \\
                        & 0.110 $\pm$ 0.022 & 0.176 $\pm$ 0.021 & 0.078 $\pm$ 0.012 & 0.222 $\pm$ 0.020 & 0.184 $\pm$ 0.009 & 0.078 $\pm$ 0.015 \\ \midrule
    ForwGAN             & 0.083 $\pm$ 0.016 & 0.159 $\pm$ 0.017 & 0.073 $\pm$ 0.008 & 0.196 $\pm$ 0.021 & 0.233 $\pm$ 0.020 & 0.107 $\pm$ 0.019 \\
                        & 0.106 $\pm$ 0.021 & 0.179 $\pm$ 0.020 & 0.075 $\pm$ 0.013 & 0.210 $\pm$ 0.021 & 0.187 $\pm$ 0.011 & 0.098 $\pm$ 0.022 \\
                        & 0.096 $\pm$ 0.026 & 0.181 $\pm$ 0.030 & 0.055 $\pm$ 0.011 & 0.205 $\pm$ 0.021 & 0.168 $\pm$ 0.014 & 0.075 $\pm$ 0.014 \\ \midrule
    RevGAN              & 0.097 $\pm$ 0.027 & 0.147 $\pm$ 0.022 & 0.074 $\pm$ 0.010 & 0.172 $\pm$ 0.014 & 0.206 $\pm$ 0.022 & 0.104 $\pm$ 0.021 \\
                        & 0.103 $\pm$ 0.025 & 0.173 $\pm$ 0.021 & 0.080 $\pm$ 0.009 & 0.188 $\pm$ 0.018 & 0.197 $\pm$ 0.013 & 0.087 $\pm$ 0.019 \\
                        & 0.101 $\pm$ 0.015 & 0.174 $\pm$ 0.016 & 0.062 $\pm$ 0.011 & 0.187 $\pm$ 0.019 & 0.185 $\pm$ 0.015 & 0.085 $\pm$ 0.019 \\ \midrule
    JSGAN               & 0.245 $\pm$ 0.034 & 0.183 $\pm$ 0.016 & 0.116 $\pm$ 0.020 & 0.352 $\pm$ 0.036 & 0.229 $\pm$ 0.020 & 0.147 $\pm$ 0.026 \\
                        & 0.243 $\pm$ 0.035 & 0.175 $\pm$ 0.013 & 0.124 $\pm$ 0.019 & 0.324 $\pm$ 0.025 & 0.195 $\pm$ 0.046 & 0.150 $\pm$ 0.034 \\
                        & 0.257 $\pm$ 0.038 & 0.187 $\pm$ 0.012 & 0.097 $\pm$ 0.018 & 0.339 $\pm$ 0.055 & 0.167 $\pm$ 0.026 & 0.086 $\pm$ 0.016 \\ \midrule
    TVGAN               & 0.095 $\pm$ 0.022 & 0.165 $\pm$ 0.018 & 0.081 $\pm$ 0.014 & 0.204 $\pm$ 0.024 & 0.215 $\pm$ 0.022 & 0.126 $\pm$ 0.026 \\
                        & 0.098 $\pm$ 0.016 & 0.157 $\pm$ 0.012 & 0.079 $\pm$ 0.009 & 0.205 $\pm$ 0.025 & 0.188 $\pm$ 0.013 & 0.123 $\pm$ 0.024 \\ 
                        & 0.101 $\pm$ 0.016 & 0.183 $\pm$ 0.016 & 0.085 $\pm$ 0.013 & 0.204 $\pm$ 0.035 & 0.176 $\pm$ 0.012 & 0.062 $\pm$ 0.011 \\ \midrule
    HellingerGAN        & 0.088 $\pm$ 0.021 & 0.148 $\pm$ 0.014 & 0.067 $\pm$ 0.011 & 0.167 $\pm$ 0.012 & 0.215 $\pm$ 0.024 & 0.102 $\pm$ 0.022 \\
                        & 0.097 $\pm$ 0.018 & 0.152 $\pm$ 0.013 & 0.072 $\pm$ 0.015 & 0.196 $\pm$ 0.014 & 0.210 $\pm$ 0.012 & 0.104 $\pm$ 0.018 \\
                        & 0.099 $\pm$ 0.022 & 0.159 $\pm$ 0.018 & 0.062 $\pm$ 0.016 & 0.179 $\pm$ 0.021 & 0.212 $\pm$ 0.022 & 0.076 $\pm$ 0.014 \\ \midrule
    PearsonGAN          & 0.112 $\pm$ 0.023 & 0.172 $\pm$ 0.021 & 0.087 $\pm$ 0.013 & 0.238 $\pm$ 0.024 & 0.217 $\pm$ 0.024 & 0.124 $\pm$ 0.023 \\
                        & 0.118 $\pm$ 0.021 & 0.177 $\pm$ 0.014 & 0.088 $\pm$ 0.008 & 0.256 $\pm$ 0.027 & 0.203 $\pm$ 0.018 & 0.121 $\pm$ 0.026 \\
                        & 0.100 $\pm$ 0.013 & 0.180 $\pm$ 0.020 & 0.084 $\pm$ 0.009 & 0.250 $\pm$ 0.030 & 0.175 $\pm$ 0.016 & 0.074 $\pm$ 0.015 \\ \midrule
    \end{tabular}}
  \caption{Confidence intervals of the best average minimum Jensen-Shannon divergence across 20 trials for 1,000 samples (first row), 10,000 samples (second row), and 100,000 samples (third row) of the best hyperparameter settings for dimensionality $N=128$.} \label{tab:jsconfidenceintervals}
\end{table*}

\begin{table*}[!htb]
  \centering
  \resizebox{\textwidth}{!}{%
    \begin{tabular}{lllllll}
    Model                & Normal                       & Beta         & Gumbel       & Laplace      & Exponential   &               Gamma               \\ \toprule
    WGAN                 & 0.832 $\pm$ 0.295 & 0.445 $\pm$ 0.043 & 0.251 $\pm$ 0.037 & 1.174 $\pm$ 0.249 & 0.609 $\pm$ 0.061 & 0.312 $\pm$ 0.038 \\
                         & 0.985 $\pm$ 0.188 & 0.521 $\pm$ 0.082 & 0.232 $\pm$ 0.025 & 1.359 $\pm$ 0.221 & 0.639 $\pm$ 0.052 & 0.229 $\pm$ 0.037 \\
                         & 1.185 $\pm$ 0.194 & 0.566 $\pm$ 0.083 & 0.216 $\pm$ 0.025 & 1.430 $\pm$ 0.285 & 0.552 $\pm$ 0.040 & 0.244 $\pm$ 0.040 \\ \midrule
    WGANGP               & 0.989 $\pm$ 0.280 & 0.985 $\pm$ 0.119 & 1.139 $\pm$ 0.303 & 1.916 $\pm$ 0.351 & 0.955 $\pm$ 0.122 & 0.904 $\pm$ 0.144 \\
                         & 1.520 $\pm$ 0.430 & 0.610 $\pm$ 0.091 & 0.440 $\pm$ 0.085 & 1.943 $\pm$ 0.412 & 0.661 $\pm$ 0.142 & 0.393 $\pm$ 0.177 \\
                         & 1.380 $\pm$ 0.134 & 0.401 $\pm$ 0.131 & 0.318 $\pm$ 0.041 & 1.533 $\pm$ 0.184 & 0.463 $\pm$ 0.042 & 0.216 $\pm$ 0.042 \\ \midrule
    NSGAN                & 0.544 $\pm$ 0.238 & 0.814 $\pm$ 0.256 & 0.177 $\pm$ 0.010 & 0.891 $\pm$ 0.202 & 0.687 $\pm$ 0.206 & 0.603 $\pm$ 0.299 \\
                         & 0.381 $\pm$ 0.181 & 0.583 $\pm$ 0.080 & 0.221 $\pm$ 0.075 & 0.874 $\pm$ 0.256 & 0.546 $\pm$ 0.170 & 0.534 $\pm$ 0.204 \\
                         & 0.522 $\pm$ 0.179 & 0.473 $\pm$ 0.064 & 0.165 $\pm$ 0.011 & 0.788 $\pm$ 0.249 & 0.556 $\pm$ 0.078 & 0.302 $\pm$ 0.069 \\ \midrule
    MMGAN                & 2.458 $\pm$ 0.184 & 0.932 $\pm$ 0.114 & 0.864 $\pm$ 0.228 & 2.796 $\pm$ 0.373 & 1.028 $\pm$ 0.115 & 1.108 $\pm$ 0.158 \\
                         & 1.946 $\pm$ 0.219 & 0.697 $\pm$ 0.094 & 0.374 $\pm$ 0.041 & 2.290 $\pm$ 0.347 & 0.999 $\pm$ 0.146 & 0.584 $\pm$ 0.185 \\
                         & 1.449 $\pm$ 0.327 & 0.572 $\pm$ 0.127 & 0.310 $\pm$ 0.076 & 2.013 $\pm$ 0.155 & 0.598 $\pm$ 0.086 & 0.254 $\pm$ 0.038 \\ \midrule
    RaGAN                & 0.730 $\pm$ 0.299 & 0.992 $\pm$ 0.126 & 0.211 $\pm$ 0.077 & 0.900 $\pm$ 0.187 & 0.940 $\pm$ 0.101 & 0.935 $\pm$ 0.368 \\
                         & 0.795 $\pm$ 0.342 & 1.067 $\pm$ 0.129 & 0.262 $\pm$ 0.186 & 1.167 $\pm$ 0.323 & 1.034 $\pm$ 0.126 & 0.451 $\pm$ 0.160 \\
                         & 0.741 $\pm$ 0.242 & 1.183 $\pm$ 0.337 & 0.221 $\pm$ 0.080 & 1.040 $\pm$ 0.243 & 1.052 $\pm$ 0.126 & 0.357 $\pm$ 0.085 \\ \midrule
    LSGAN                & 1.012 $\pm$ 0.369 & 0.901 $\pm$ 0.155 & 0.616 $\pm$ 0.157 & 1.006 $\pm$ 0.196 & 0.950 $\pm$ 0.062 & 1.035 $\pm$ 0.118 \\
                         & 1.608 $\pm$ 0.316 & 0.633 $\pm$ 0.103 & 0.390 $\pm$ 0.112 & 2.004 $\pm$ 0.324 & 0.925 $\pm$ 0.221 & 0.675 $\pm$ 0.257 \\
                         & 1.073 $\pm$ 0.102 & 0.365 $\pm$ 0.128 & 0.316 $\pm$ 0.038 & 1.728 $\pm$ 0.221 & 0.495 $\pm$ 0.070 & 0.274 $\pm$ 0.075 \\ \midrule
    DRAGAN               & 2.326 $\pm$ 0.230 & 1.075 $\pm$ 0.153 & 1.281 $\pm$ 0.120 & 3.223 $\pm$ 0.270 & 0.930 $\pm$ 0.089 & 0.993 $\pm$ 0.094 \\
                         & 2.444 $\pm$ 0.213 & 1.367 $\pm$ 0.187 & 1.314 $\pm$ 0.162 & 3.348 $\pm$ 0.301 & 1.008 $\pm$ 0.144 & 1.001 $\pm$ 0.092 \\
                         & 1.608 $\pm$ 0.226 & 0.683 $\pm$ 0.192 & 0.594 $\pm$ 0.139 & 2.119 $\pm$ 0.376 & 0.732 $\pm$ 0.286 & 0.879 $\pm$ 0.326 \\ \midrule
    BEGAN                & 2.774 $\pm$ 0.268 & 1.360 $\pm$ 0.118 & 1.445 $\pm$ 0.170 & 3.721 $\pm$ 0.347 & 1.060 $\pm$ 0.129 & 1.144 $\pm$ 0.131 \\
                         & 3.004 $\pm$ 0.339 & 1.394 $\pm$ 0.147 & 1.008 $\pm$ 0.452 & 2.389 $\pm$ 0.421 & 1.269 $\pm$ 0.149 & 1.391 $\pm$ 0.217 \\
                         & 2.099 $\pm$ 0.368 & 1.446 $\pm$ 0.134 & 0.589 $\pm$ 0.183 & 1.834 $\pm$ 0.421 & 1.645 $\pm$ 0.257 & 1.209 $\pm$ 0.316 \\ \midrule
    InfoGAN              & 1.021 $\pm$ 0.187 & 0.510 $\pm$ 0.103 & 0.561 $\pm$ 0.121 & 1.464 $\pm$ 0.220 & 0.826 $\pm$ 0.118 & 0.664 $\pm$ 0.208 \\
                         & 0.882 $\pm$ 0.158 & 0.505 $\pm$ 0.115 & 0.371 $\pm$ 0.024 & 0.906 $\pm$ 0.110 & 0.593 $\pm$ 0.081 & 0.336 $\pm$ 0.051 \\
                         & 0.965 $\pm$ 0.197 & 0.428 $\pm$ 0.109 & 0.370 $\pm$ 0.022 & 1.213 $\pm$ 0.076 & 0.456 $\pm$ 0.051 & 0.321 $\pm$ 0.044 \\ \midrule
    FisherGAN            & 0.391 $\pm$ 0.173 & 0.709 $\pm$ 0.060 & 0.171 $\pm$ 0.015 & 0.779 $\pm$ 0.240 & 0.803 $\pm$ 0.290 & 0.381 $\pm$ 0.149 \\
                         & 0.609 $\pm$ 0.181 & 0.818 $\pm$ 0.061 & 0.235 $\pm$ 0.081 & 0.906 $\pm$ 0.182 & 0.621 $\pm$ 0.202 & 0.394 $\pm$ 0.180 \\
                         & 0.483 $\pm$ 0.172 & 0.646 $\pm$ 0.166 & 0.176 $\pm$ 0.013 & 0.637 $\pm$ 0.129 & 0.716 $\pm$ 0.224 & 0.346 $\pm$ 0.100 \\ \midrule
    ForwGAN              & 0.321 $\pm$ 0.137 & 0.680 $\pm$ 0.202 & 0.190 $\pm$ 0.009 & 0.771 $\pm$ 0.219 & 0.909 $\pm$ 0.218 & 0.397 $\pm$ 0.134 \\
                         & 0.377 $\pm$ 0.138 & 0.808 $\pm$ 0.160 & 0.232 $\pm$ 0.151 & 0.653 $\pm$ 0.150 & 0.873 $\pm$ 0.210 & 0.466 $\pm$ 0.280 \\
                         & 0.541 $\pm$ 0.310 & 0.766 $\pm$ 0.219 & 0.159 $\pm$ 0.014 & 0.735 $\pm$ 0.265 & 0.712 $\pm$ 0.150 & 0.334 $\pm$ 0.087 \\ \midrule
    RevGAN               & 0.518 $\pm$ 0.174 & 0.690 $\pm$ 0.133 & 0.191 $\pm$ 0.074 & 0.698 $\pm$ 0.195 & 0.820 $\pm$ 0.188 & 0.352 $\pm$ 0.132 \\
                         & 0.519 $\pm$ 0.211 & 0.745 $\pm$ 0.197 & 0.214 $\pm$ 0.019 & 0.741 $\pm$ 0.208 & 0.779 $\pm$ 0.222 & 0.369 $\pm$ 0.151 \\
                         & 0.370 $\pm$ 0.170 & 0.630 $\pm$ 0.175 & 0.150 $\pm$ 0.013 & 0.691 $\pm$ 0.135 & 0.863 $\pm$ 0.210 & 0.400 $\pm$ 0.137 \\ \midrule
    JSGAN                & 1.919 $\pm$ 0.639 & 1.014 $\pm$ 0.115 & 0.219 $\pm$ 0.071 & 1.705 $\pm$ 0.445 & 0.905 $\pm$ 0.087 & 1.050 $\pm$ 0.137 \\
                         & 1.560 $\pm$ 0.439 & 0.904 $\pm$ 0.175 & 0.521 $\pm$ 0.217 & 1.855 $\pm$ 0.491 & 0.998 $\pm$ 0.130 & 0.687 $\pm$ 0.136 \\
                         & 1.879 $\pm$ 0.384 & 1.034 $\pm$ 0.226 & 0.346 $\pm$ 0.130 & 2.077 $\pm$ 0.561 & 0.858 $\pm$ 0.201 & 0.359 $\pm$ 0.083 \\ \midrule
    TVGAN                & 0.486 $\pm$ 0.188 & 0.762 $\pm$ 0.244 & 0.201 $\pm$ 0.075 & 0.911 $\pm$ 0.273 & 0.694 $\pm$ 0.242 & 0.590 $\pm$ 0.237 \\
                         & 0.602 $\pm$ 0.187 & 0.800 $\pm$ 0.248 & 0.197 $\pm$ 0.079 & 0.962 $\pm$ 0.240 & 0.798 $\pm$ 0.214 & 0.581 $\pm$ 0.220 \\
                         & 0.460 $\pm$ 0.189 & 0.670 $\pm$ 0.181 & 0.180 $\pm$ 0.015 & 0.752 $\pm$ 0.169 & 0.594 $\pm$ 0.200 & 0.273 $\pm$ 0.058 \\ \midrule
    HellingerGAN         & 0.377 $\pm$ 0.208 & 0.548 $\pm$ 0.112 & 0.155 $\pm$ 0.012 & 0.645 $\pm$ 0.195 & 0.801 $\pm$ 0.170 & 0.337 $\pm$ 0.103 \\
                         & 0.409 $\pm$ 0.190 & 0.678 $\pm$ 0.195 & 0.187 $\pm$ 0.026 & 0.628 $\pm$ 0.148 & 0.804 $\pm$ 0.238 & 0.379 $\pm$ 0.111 \\
                         & 0.352 $\pm$ 0.100 & 0.659 $\pm$ 0.200 & 0.149 $\pm$ 0.010 & 0.586 $\pm$ 0.191 & 0.855 $\pm$ 0.152 & 0.337 $\pm$ 0.076 \\ \midrule
    PearsonGAN           & 0.764 $\pm$ 0.242 & 0.869 $\pm$ 0.216 & 0.177 $\pm$ 0.015 & 0.874 $\pm$ 0.286 & 0.578 $\pm$ 0.147 & 0.409 $\pm$ 0.202 \\
                         & 0.600 $\pm$ 0.229 & 0.815 $\pm$ 0.202 & 0.175 $\pm$ 0.025 & 0.862 $\pm$ 0.415 & 0.786 $\pm$ 0.214 & 0.744 $\pm$ 0.169 \\
                         & 0.391 $\pm$ 0.143 & 0.867 $\pm$ 0.262 & 0.154 $\pm$ 0.017 & 0.820 $\pm$ 0.246 & 0.481 $\pm$ 0.097 & 0.318 $\pm$ 0.099 \\ \bottomrule
    \end{tabular}}
  \caption{Confidence intervals of the best average minimum Kullback-Leibler divergence across 20 trials for 1,000 samples (first row), 10,000 samples (second row), and 100,000 samples (third row) of the best hyperparameter settings for dimensionality $N=128$.} \label{tab:klconfidenceintervals}
\end{table*}

\begin{table*}[!htb]
  \centering
  \resizebox{\textwidth}{!}{%
    \begin{tabular}{lllllll}
    Model                & Normal                       & Beta         & Gumbel       & Laplace      & Exponential   &               Gamma          \\ \toprule
    WGAN              & 2.376 $\pm$ 0.428  & 6.301 $\pm$ 0.543  & 5.910 $\pm$ 1.368  & 7.698 $\pm$ 2.711  & 9.615 $\pm$ 2.024  & 3.235 $\pm$ 1.053  \\
                      & 2.655 $\pm$ 0.950  & 6.777 $\pm$ 2.090  & 8.019 $\pm$ 1.944  & 8.577 $\pm$ 3.843  & 11.899 $\pm$ 5.086 & 2.995 $\pm$ 1.149  \\
                      & 3.042 $\pm$ 0.449  & 5.927 $\pm$ 0.366  & 7.037 $\pm$ 2.640  & 6.988 $\pm$ 1.975  & 8.405 $\pm$ 1.402  & 1.881 $\pm$ 0.543  \\ \midrule
    WGANGP            & 3.191 $\pm$ 0.400  & 8.336 $\pm$ 0.654  & 7.611 $\pm$ 2.508  & 8.480 $\pm$ 3.452  & 11.127 $\pm$ 2.078 & 4.279 $\pm$ 1.734  \\
                      & 4.195 $\pm$ 1.431  & 7.808 $\pm$ 2.472  & 9.871 $\pm$ 4.512  & 9.882 $\pm$ 5.517  & 11.367 $\pm$ 6.599 & 3.672 $\pm$ 1.154  \\
                      & 2.043 $\pm$ 0.147  & 5.038 $\pm$ 0.290  & 4.768 $\pm$ 1.626  & 4.941 $\pm$ 1.163  & 8.142 $\pm$ 3.307  & 1.348 $\pm$ 0.396  \\ \midrule
    NSGAN             & 1.749 $\pm$ 0.413  & 9.099 $\pm$ 0.734  & 7.445 $\pm$ 2.463  & 6.187 $\pm$ 1.414  & 10.076 $\pm$ 3.777 & 3.401 $\pm$ 1.031  \\
                      & 3.358 $\pm$ 1.237  & 8.643 $\pm$ 2.142  & 8.464 $\pm$ 2.808  & 8.950 $\pm$ 3.988  & 11.447 $\pm$ 4.605 & 3.737 $\pm$ 1.477  \\
                      & 3.029 $\pm$ 0.229  & 7.469 $\pm$ 0.535  & 4.117 $\pm$ 1.023  & 5.327 $\pm$ 1.703  & 10.218 $\pm$ 3.821 & 2.223 $\pm$ 0.816  \\ \midrule
    MMGAN             & 10.201 $\pm$ 1.561 & 10.464 $\pm$ 0.755 & 11.390 $\pm$ 2.192 & 13.033 $\pm$ 1.864 & 17.059 $\pm$ 3.348 & 10.312 $\pm$ 1.613 \\
                      & 5.495 $\pm$ 1.363  & 10.551 $\pm$ 3.462 & 8.330 $\pm$ 3.214  & 10.662 $\pm$ 5.150 & 18.233 $\pm$ 6.909 & 5.926 $\pm$ 5.555  \\
                      & 3.110 $\pm$ 0.452  & 6.617 $\pm$ 0.661  & 4.452 $\pm$ 1.029  & 5.534 $\pm$ 1.823  & 10.873 $\pm$ 2.759 & 1.843 $\pm$ 0.506  \\ \midrule
    RaGAN             & 2.740 $\pm$ 0.566  & 8.983 $\pm$ 0.788  & 7.380 $\pm$ 2.084  & 7.509 $\pm$ 2.185  & 8.976 $\pm$ 2.146  & 2.791 $\pm$ 1.016  \\
                      & 4.519 $\pm$ 1.452  & 13.032 $\pm$ 4.097 & 10.823 $\pm$ 4.832 & 8.052 $\pm$ 2.342  & 12.526 $\pm$ 3.182 & 4.056 $\pm$ 1.408  \\
                      & 3.537 $\pm$ 0.381  & 10.291 $\pm$ 0.733 & 7.178 $\pm$ 1.257  & 7.852 $\pm$ 2.715  & 10.936 $\pm$ 2.011 & 3.654 $\pm$ 1.024  \\ \midrule
    LSGAN             & 4.280 $\pm$ 0.518  & 9.232 $\pm$ 0.563  & 10.503 $\pm$ 3.403 & 8.753 $\pm$ 2.167  & 11.754 $\pm$ 3.206 & 9.023 $\pm$ 0.929  \\
                      & 4.096 $\pm$ 1.636  & 8.427 $\pm$ 2.414  & 8.978 $\pm$ 4.606  & 13.346 $\pm$ 4.851 & 15.695 $\pm$ 5.691 & 7.326 $\pm$ 5.143  \\
                      & 2.286 $\pm$ 0.190  & 5.827 $\pm$ 0.778  & 3.460 $\pm$ 0.814  & 4.262 $\pm$ 0.798  & 9.539 $\pm$ 3.441  & 2.345 $\pm$ 0.962  \\ \midrule
    DRAGAN            & 7.051 $\pm$ 0.851  & 8.892 $\pm$ 0.572  & 11.208 $\pm$ 1.052 & 11.004 $\pm$ 1.971 & 15.461 $\pm$ 2.439 & 10.695 $\pm$ 2.428 \\
                      & 9.638 $\pm$ 2.113  & 11.497 $\pm$ 4.364 & 17.183 $\pm$ 6.580 & 12.998 $\pm$ 3.925 & 19.594 $\pm$ 9.288 & 9.541 $\pm$ 5.206  \\
                      & 2.704 $\pm$ 0.338  & 4.614 $\pm$ 0.641  & 5.762 $\pm$ 2.537  & 6.923 $\pm$ 1.181  & 7.696 $\pm$ 2.245  & 2.713 $\pm$ 0.689  \\ \midrule
    BEGAN             & 14.073 $\pm$ 0.759 & 11.215 $\pm$ 0.477 & 20.592 $\pm$ 2.197 & 18.819 $\pm$ 2.514 & 21.311 $\pm$ 4.300 & 16.569 $\pm$ 0.759 \\
                      & 19.480 $\pm$ 5.667 & 12.881 $\pm$ 3.939 & 23.892 $\pm$ 7.131 & 22.189 $\pm$ 6.844 & 24.741 $\pm$ 7.144 & 14.040 $\pm$ 9.001 \\
                      & 13.694 $\pm$ 1.408 & 10.468 $\pm$ 0.663 & 16.832 $\pm$ 3.186 & 20.066 $\pm$ 1.192 & 23.807 $\pm$ 3.008 & 9.741 $\pm$ 1.698  \\ \midrule
    InfoGAN           & 4.184 $\pm$ 0.693  & 7.700 $\pm$ 0.388  & 7.702 $\pm$ 2.137  & 8.213 $\pm$ 2.313  & 10.682 $\pm$ 1.933 & 5.653 $\pm$ 1.414  \\
                      & 4.707 $\pm$ 3.152  & 8.039 $\pm$ 3.353  & 8.623 $\pm$ 3.280  & 8.085 $\pm$ 3.983  & 12.106 $\pm$ 8.355 & 5.034 $\pm$ 3.126  \\
                      & 2.535 $\pm$ 0.277  & 5.789 $\pm$ 0.597  & 6.710 $\pm$ 2.561  & 6.351 $\pm$ 2.116  & 9.892 $\pm$ 3.821  & 2.207 $\pm$ 0.638  \\ \midrule
    FisherGAN         & 1.823 $\pm$ 0.550  & 8.931 $\pm$ 0.997  & 5.493 $\pm$ 1.880  & 6.836 $\pm$ 1.186  & 10.790 $\pm$ 3.162 & 3.579 $\pm$ 2.350  \\
                      & 3.105 $\pm$ 1.345  & 11.792 $\pm$ 3.672 & 7.582 $\pm$ 2.777  & 8.348 $\pm$ 4.066  & 12.528 $\pm$ 3.745 & 3.895 $\pm$ 1.670  \\
                      & 2.768 $\pm$ 0.433  & 9.146 $\pm$ 0.449  & 7.666 $\pm$ 1.595  & 5.885 $\pm$ 0.885  & 10.648 $\pm$ 2.411 & 3.388 $\pm$ 0.929  \\ \midrule
    ForwGAN           & 1.450 $\pm$ 0.426  & 8.835 $\pm$ 0.506  & 7.515 $\pm$ 2.029  & 5.828 $\pm$ 1.839  & 11.519 $\pm$ 3.067 & 3.431 $\pm$ 0.767  \\
                      & 2.634 $\pm$ 0.958  & 11.460 $\pm$ 3.812 & 9.553 $\pm$ 4.423  & 7.038 $\pm$ 2.696  & 11.161 $\pm$ 2.903 & 4.347 $\pm$ 1.576  \\
                      & 2.380 $\pm$ 0.413  & 9.478 $\pm$ 0.281  & 6.854 $\pm$ 2.344  & 4.799 $\pm$ 0.825  & 8.786 $\pm$ 1.829  & 3.220 $\pm$ 0.977  \\ \midrule
    RevGAN            & 1.537 $\pm$ 0.456  & 8.915 $\pm$ 0.627  & 6.060 $\pm$ 1.804  & 4.752 $\pm$ 0.978  & 10.211 $\pm$ 3.843 & 2.889 $\pm$ 1.359  \\
                      & 2.358 $\pm$ 0.593  & 11.499 $\pm$ 3.694 & 9.208 $\pm$ 3.809  & 6.812 $\pm$ 2.425  & 10.693 $\pm$ 3.902 & 3.420 $\pm$ 1.207  \\
                      & 2.714 $\pm$ 0.336  & 9.645 $\pm$ 0.424  & 5.959 $\pm$ 1.767  & 5.355 $\pm$ 1.093  & 7.898 $\pm$ 1.725  & 3.665 $\pm$ 1.149  \\ \midrule
    JSGAN             & 5.042 $\pm$ 0.709  & 10.094 $\pm$ 0.601 & 10.782 $\pm$ 2.684 & 9.379 $\pm$ 2.631  & 12.464 $\pm$ 1.795 & 4.090 $\pm$ 1.270  \\
                      & 5.276 $\pm$ 1.471  & 11.595 $\pm$ 2.963 & 12.447 $\pm$ 5.515 & 10.820 $\pm$ 2.957 & 14.465 $\pm$ 5.005 & 5.057 $\pm$ 2.490  \\
                      & 4.272 $\pm$ 0.908  & 10.478 $\pm$ 1.770 & 8.104 $\pm$ 1.919  & 9.201 $\pm$ 1.930  & 12.593 $\pm$ 2.252 & 3.932 $\pm$ 0.745  \\ \midrule
    TVGAN             & 1.572 $\pm$ 0.425  & 9.026 $\pm$ 0.685  & 7.761 $\pm$ 1.566  & 6.326 $\pm$ 0.902  & 8.706 $\pm$ 1.955  & 2.868 $\pm$ 1.068  \\
                      & 3.024 $\pm$ 1.161  & 11.396 $\pm$ 3.091 & 8.336 $\pm$ 2.476  & 6.074 $\pm$ 2.376  & 10.191 $\pm$ 3.094 & 3.592 $\pm$ 1.347  \\
                      & 2.543 $\pm$ 0.421  & 9.273 $\pm$ 0.436  & 8.183 $\pm$ 2.106  & 6.407 $\pm$ 2.109  & 8.852 $\pm$ 2.168  & 3.529 $\pm$ 0.848  \\ \midrule
    HellingerGAN      & 1.524 $\pm$ 0.341  & 8.882 $\pm$ 0.431  & 6.097 $\pm$ 1.572  & 5.846 $\pm$ 1.577  & 10.578 $\pm$ 2.485 & 3.140 $\pm$ 1.179  \\
                      & 2.694 $\pm$ 0.781  & 11.558 $\pm$ 3.325 & 8.278 $\pm$ 1.857  & 7.158 $\pm$ 2.250  & 10.792 $\pm$ 4.651 & 4.148 $\pm$ 1.731  \\
                      & 2.376 $\pm$ 0.346  & 9.982 $\pm$ 0.520  & 6.650 $\pm$ 1.888  & 6.172 $\pm$ 2.072  & 7.931 $\pm$ 1.591  & 2.925 $\pm$ 1.307  \\ \midrule
    PearsonGAN        & 1.946 $\pm$ 0.439  & 9.315 $\pm$ 0.996  & 7.482 $\pm$ 3.103  & 5.619 $\pm$ 1.029  & 8.804 $\pm$ 1.810  & 2.961 $\pm$ 0.885  \\
                      & 3.243 $\pm$ 1.122  & 11.982 $\pm$ 4.928 & 7.740 $\pm$ 2.779  & 8.301 $\pm$ 3.403  & 13.876 $\pm$ 4.201 & 4.569 $\pm$ 1.747  \\
                      & 2.639 $\pm$ 0.246  & 9.390 $\pm$ 0.469  & 7.866 $\pm$ 1.670  & 7.304 $\pm$ 1.037  & 10.162 $\pm$ 2.995 & 3.465 $\pm$ 0.858  \\ \bottomrule
    \end{tabular}}
  \caption{Confidence intervals of the best average minimum Wasserstein distance across 20 trials for 1,000 samples (first row), 10,000 samples (second row), and 100,000 samples (third row) of the best hyperparameter settings for dimensionality $N=128$.} \label{tab:wdconfidenceintervals}
\end{table*}

\clearpage

\clearpage

\section{Diverse sets of hyperparameters can produce a best result}

\begin{table*}[!htb]
  \centering
  \scriptsize
  \begin{tabular}{lllllll}
 Model               & Normal      & Beta       & Gumbel       & Laplace     & Exponential   & Gamma        \\ \toprule
 WGAN                & [8, 3, 13]  & [5, 3, 3]  & [6, 6, 7]    & [6, 5, 7]   & [8, 11, 9]    & [11, 8, 7]   \\
                     & [14, 12, 10] & [5, 3, 4]   & [5, 10, 5] & [14, 9, 9]   & [10, 9, 10]   & [11, 9, 7] \\
                     & [8, 11, 9]   & [7, 6, 7]  & [13, 11, 14] & [8, 10, 12]  & [13, 13, 13]  & [8, 10, 12] \\ \midrule
 WGANGP              & [4, 8, 9]   & [2, 3, 5]  & [8, 7, 3]    & [4, 5, 6]   & [3, 6, 5]     & [9, 4, 6]    \\
                     & [5, 7, 7]    & [3, 3, 3]   & [6, 5, 2]  & [4, 5, 4]    & [8, 8, 10]    & [11, 5, 6] \\
                     & [8, 6, 7]    & [8, 8, 8]  & [13, 13, 13] & [13, 12, 14] & [15, 12, 14]  & [11, 10, 14] \\ \midrule
 NSGAN               & [6, 5, 5]   & [7, 4, 4]  & [5, 7, 10]   & [7, 4, 5]   & [9, 4, 9]     & [11, 8, 8]   \\
                     & [9, 7, 8]    & [9, 4, 4]   & [5, 8, 7]  & [8, 5, 6]    & [9, 10, 12]   & [11, 8, 12] \\
                     & [5, 6, 7]    & [3, 5, 6]  & [11, 10, 12] & [8, 11, 14]  & [11, 9, 14]   & [10, 9, 12] \\ \midrule
 MMGAN               & [9, 8, 6]   & [5, 4, 5]  & [5, 4, 6]    & [12, 11, 5] & [4, 5, 8]     & [9, 7, 6]    \\
                     & [8, 6, 4]    & [3, 4, 5]   & [6, 4, 5]  & [7, 4, 2]    & [5, 7, 10]    & [5, 3, 7] \\
                     & [15, 8, 8]   & [7, 3, 9]  & [14, 6, 13]  & [14, 9, 15]  & [17, 11, 12]  & [14, 15, 11] \\ \midrule
 RaGAN               & [11, 8, 7]  & [4, 6, 9]  & [8, 10, 9]   & [8, 9, 7]   & [8, 5, 5]     & [13, 12, 13] \\ 
                     & [11, 12, 10] & [6, 9, 10]  & [9, 9, 13] & [13, 13, 11] & [8, 4, 4]     & [12, 9, 11] \\
                     & [9, 5, 10]   & [5, 8, 12] & [9, 14, 12]  & [12, 10, 13] & [12, 12, 10]  & [14, 12, 13] \\ \midrule
 LSGAN               & [4, 9, 9]   & [5, 5, 7]  & [3, 5, 5]    & [2, 6, 8]   & [6, 5, 5]     & [3, 7, 7]    \\
                     & [5, 4, 3]    & [5, 7, 6]   & [3, 4, 5]  & [2, 7, 4]    & [8, 7, 7]     & [4, 6, 8] \\
                     & [4, 6, 7]    & [4, 8, 8]  & [8, 10, 12]  & [6, 11, 11]  & [9, 12, 11]   & [9, 7, 12] \\ \midrule
 DRAGAN              & [12, 11, 7] & [9, 4, 5]  & [9, 9, 6]    & [9, 11, 9]  & [9, 8, 4]     & [14, 13, 7]  \\
                     & [8, 6, 9]    & [6, 4, 9]   & [5, 4, 6]  & [11, 6, 8]   & [7, 4, 11]    & [10, 7, 3] \\
                     & [13, 13, 5]  & [6, 12, 5] & [10, 10, 10] & [12, 12, 9]  & [12, 9, 9]    & [11, 13, 7] \\ \midrule
 BEGAN               & [11, 4, 5]  & [8, 5, 5]  & [7, 4, 3]    & [6, 3, 3]   & [6, 3, 6]     & [4, 5, 3]    \\
                     & [7, 3, 5]    & [6, 6, 4]   & [5, 5, 3]  & [6, 4, 2]    & [4, 3, 5]     & [4, 2, 4] \\
                     & [7, 9, 4]    & [12, 8, 9] & [17, 15, 4]  & [13, 9, 3]   & [15, 13, 15]  & [9, 9, 3] \\ \midrule
 InfoGAN             & [10, 10, 5] & [3, 4, 5]  & [9, 5, 9]    & [8, 10, 7]  & [8, 4, 2]     & [9, 4, 7]    \\
                     & [7, 5, 3]    & [4, 4, 5]   & [8, 3, 8]  & [5, 6, 5]    & [9, 3, 4]     & [12, 4, 9] \\
                     & [7, 7, 9]    & [7, 10, 3] & [12, 11, 11] & [13, 13, 10] & [13, 14, 12]  & [12, 4, 12] \\ \midrule
 FisherGAN           & [6, 7, 5]   & [9, 8, 7]  & [6, 8, 9]    & [8, 9, 7]   & [10, 9, 13]   & [10, 11, 9]  \\
                     & [9, 12, 7]   & [9, 10, 9]  & [5, 8, 6]  & [9, 9, 7]    & [10, 7, 8]    & [10, 10, 12] \\
                     & [7, 6, 10]   & [4, 6, 6]  & [12, 10, 13] & [10, 11, 10] & [11, 13, 15]  & [11, 10, 9] \\ \midrule
 ForwGAN     & [5, 6, 6]   & [7, 9, 11] & [6, 6, 6]    & [7, 6, 7]   & [4, 10, 12]   & [8, 7, 10]   \\
                      & [8, 8, 8]    & [8, 6, 7]   & [3, 6, 4]  & [9, 7, 7]    & [11, 12, 9]   & [11, 8, 10] \\
                      & [5, 6, 6]    & [6, 8, 6]  & [12, 14, 13] & [10, 11, 8]  & [11, 11, 13]  & [9, 11, 10] \\ \midrule
 RevGAN     & [5, 8, 9]   & [5, 10, 7] & [7, 7, 8]    & [5, 5, 5]   & [6, 11, 9]    & [9, 8, 8]    \\
                      & [8, 9, 9]    & [9, 5, 7]   & [4, 5, 3]  & [7, 8, 7]    & [11, 8, 10]   & [9, 13, 12] \\
                      & [7, 7, 8]    & [7, 8, 6]  & [11, 11, 13] & [10, 11, 10] & [12, 10, 14]  & [11, 12, 9] \\ \midrule
 JSGAN & [9, 11, 11] & [6, 6, 7]  & [10, 11, 11] & [9, 7, 7]   & [7, 5, 7]     & [11, 6, 12]  \\
                      & [6, 10, 8]   & [5, 5, 5]   & [9, 9, 9]  & [9, 9, 8]    & [6, 4, 10]    & [12, 10, 13] \\
                      & [13, 12, 10] & [4, 9, 11] & [12, 10, 12] & [12, 14, 16] & [12, 11, 11]  & [12, 10, 13] \\ \midrule
 TVGAN      & [7, 6, 7]   & [7, 7, 9]  & [7, 8, 9]    & [9, 8, 8]   & [10, 11, 11]  & [10, 11, 9]  \\
                      & [9, 11, 12]  & [10, 10, 9] & [9, 6, 5]  & [9, 6, 9]    & [7, 9, 10]    & [11, 12, 11] \\
                      & [7, 7, 7]    & [5, 5, 7]  & [11, 10, 15] & [9, 9, 9]    & [12, 12, 10]  & [11, 8, 11] \\ \midrule
 HellingerGAN      & [7, 5, 5]   & [7, 8, 8]  & [6, 6, 4]    & [4, 4, 5]   & [7, 8, 6]     & [11, 9, 11]  \\
                      & [9, 6, 9]    & [6, 6, 4]   & [5, 4, 3]  & [7, 5, 6]    & [9, 6, 6]     & [11, 8, 8] \\
                      & [7, 6, 8]    & [5, 5, 9]  & [11, 10, 13] & [10, 11, 9]  & [12, 14, 14]  & [10, 11, 9] \\ \midrule
 PearsonGAN        & [9, 9, 8]   & [6, 8, 8]  & [6, 7, 6]    & [6, 8, 7]   & [8, 6, 10]    & [11, 9, 11]  \\
                      & [13, 9, 10]  & [8, 9, 8]   & [8, 8, 10] & [11, 13, 11] & [10, 7, 10]   & [14, 10, 11] \\
                      & [7, 6, 9]    & [6, 11, 6] & [11, 9, 13]  & [10, 11, 12] & [9, 12, 16]   & [11, 11, 12] \\ \bottomrule
\hline
\end{tabular}

  \caption{Number of unique hyperparameter settings that yielded a best result out of 20 trials for 1,000 samples (first entry of each row), 10,000 samples (second entry), and 100,000 samples (third entry) with respect to Kullback-Leibler divergence (first row), Jensen-Shannon divergence (second row), and Wasserstein distance (third row) for dimensionality $N=128$. The maximum value for any given entry would be 20 provided that a diferent hyperparameter output the minimum performance for every trial conducted, and the minimum value would be 1 if only one hyperparameter setting always outperformed the rest. Dimensionalities $N \in \left\{16, 32, 64 \right\}$ yielded similar results.} \label{tab:klminhyperparams}
\end{table*}

\clearpage

\section{Some GANs are more robust to hyperparameter changes than others}

\begin{table*}[!htb]
  \centering
    \begin{tabular}{lccccc}
     Model  & Jensen-Shannon  & Kullback-Leibler & Wasserstein Distance & Total (\% of max) \\ \toprule
     WGAN & 167 & 215 & 346 & 728 (1.40\%) \\ \midrule
     WGANGP & 115 & 146 & 492 & 753 (1.45\%) \\ \midrule
     NSGAN & 43 & 64 & 221 & 328 (0.63\%) \\ \midrule
     MMGAN & 34 & 127 & 416 & 577 (1.11\%) \\ \midrule
     RaGAN & 80 & 58 & 69 & 207 (0.40\%) \\ \midrule
     LSGAN & 73 & 120 & 386 & 579 (1.12\%) \\ \midrule
     DRAGAN & 126 & 138 & 380 & 644 (1.24\%) \\ \midrule
     BEGAN & 54 & 137 & 352 & 543 (1.05\%) \\ \midrule
     InfoGAN & 127 & 179 & 366 & 672 (1.30\%) \\ \midrule
     FisherGAN & 42 & 38 & 159 & 239 (0.46\%) \\ \midrule
     ForwGAN & 45 & 51 & 166 & 262 (0.51\%) \\ \midrule
     RevGAN & 43 & 44 & 137 & 224 (0.43\%)) \\ \midrule
     JSGAN & 72 & 31 & 179 & 282 (0.54\%) \\ \midrule
     TVGAN & 38 & 47 & 126 & 211 (0.41\%) \\ \midrule
     HellingerGAN & 41 & 57 & 176 & 274 (0.53\%) \\ \midrule
     PearsonGAN & 32 & 42 & 165 & 239 (0.46\%) \\ \bottomrule
    \end{tabular}
  \caption{The number of hyperparameter settings that yielded a minimum average performance that fell within the confidence interval of the best average performance across all 16 models, three evaluation metrics, six distributions, 15 hyperparameter settings, four dimensionalities, and three sample sizes. The last column indicates the sum of the values to its left followed by this number divided by 51,840 (the maximum possible number in this table) to its right, rounded up to the nearest hundredth.} \label{tab:robust}
\end{table*}

\section{Model specifications}

\begin{table*}[!htb]
  \resizebox{\textwidth}{!}{%
  \centering
    \begin{tabular}{ccccc}
        & $d=16$  & $d=32$  & $d=64$ & $d=128$  \\ \toprule
     $h=32$ & [1,393 - 1,821 - 1,888] & [2,433 - 2,861 - 3,456] & [4,513 - 4,941 - 6,592] & [8,673 - 9,101 - 12,864]  \\ \midrule
     $h=64$ & [3,281 - 4,125 - 4,256] & [5,345 - 6,189 - 7,360] & [9,473 - 10,317 -13,568] & [17,729 - 18,573 - 25,984] \\ \midrule
     $h=128$ & [8,593 - 10,269 - 10,528] & [12,705 - 14,381 - 16,704] & [20,929 - 22,605 - 29,056] & [37,377 - 39,053 - 53,760] \\ \midrule
     $h=256$ & [25,361 - 28,701 - 29,216] & [33,569 - 36,909 - 41,536] & [49,985 - 53,325 - 66,176] & [82,817 - 86,157 - 115,456] \\ \midrule
     $h=512$ & [83,473 - 90,141 - 91,168] & [99,873 - 106,541 - 115,776] & [132,673 - 139,341 - 164,992] & [198,273 - 204,941 - 263,424] \\ \bottomrule
    \end{tabular}%
  }
  \caption{Number of model parameters for data dimensionality $d$ and hidden dimension size $h$. The first entries are for all models except InfoGAN and BEGAN, the second entries are for InfoGAN, and the third are for BEGAN. All architectures consisted of four feedforward neural network layers in total: two for $G$ and two for $D$. Since InfoGAN used latent variables as inputs to $D$ and BEGAN's $D$ was an autoencoder, they necessitated slightly more parameters. We did not observe these differences to give neither InfoGAN nor BEGAN any significant advantage over other models.} \label{tab:modelparams}
\end{table*}


\begin{thebibliography}{26}
\providecommand{\natexlab}[1]{#1}
\providecommand{\url}[1]{\texttt{#1}}
\expandafter\ifx\csname urlstyle\endcsname\relax
  \providecommand{\doi}[1]{doi: #1}\else
  \providecommand{\doi}{doi: \begingroup \urlstyle{rm}\Url}\fi

\bibitem[Arjovsky et~al.(2017)Arjovsky, Chintala, and
  Bottou]{arjovsky2017wasserstein}
Mart{\'i}n Arjovsky, Soumith Chintala, and L{\'e}on Bottou.
\newblock Wasserstein generative adversarial networks.
\newblock In \emph{International Conference on Machine Learning}, pages
  214--223, 2017.

\bibitem[Arora et~al.(2017)Arora, Ge, Liang, Ma, and
  Zhang]{arora2017generalization}
Sanjeev Arora, Rong Ge, Yingyu Liang, Tengyu Ma, and Yi~Zhang.
\newblock Generalization and equilibrium in generative adversarial nets
  ({GAN}s).
\newblock In \emph{Proceedings of the 34th International Conference on Machine
  Learning}, pages 224--232, 2017.

\bibitem[Bethelot et~al.(2017)Bethelot, Schumm, and Metz]{berthelot2017began}
David Bethelot, Thomas Schumm, and Luke Metz.
\newblock {BEGAN}: Boundary equilibrium generative adversarial networks.
\newblock \emph{arXiv preprint arXiv:1703.10717}, 2017.

\bibitem[Borji(2018)]{borji2018pros}
Ali Borji.
\newblock Pros and cons of {GAN} evaluation measures.
\newblock \emph{CoRR}, abs/1802.03446, 2018.
\newblock URL \url{http://arxiv.org/abs/1802.03446}.

\bibitem[Chen et~al.(2016)Chen, Duan, Houthooft, Schulman, Sutskever, and
  Abbeel]{infogan}
Xi~Chen, Yan Duan, Rein Houthooft, John Schulman, Ilya Sutskever, and Pieter
  Abbeel.
\newblock Info{GAN}: {I}nterpretable representation learning by information
  maximizing generative adversarial nets.
\newblock In \emph{Advances in Neural Information Processing Systems}, pages
  2172--2180, 2016.

\bibitem[Endres and Schindelin(2003)]{jensenshannon}
Dominik Endres and Johannes Schindelin.
\newblock A new metric for probability distributions.
\newblock In \emph{IEEE Transactions on Information Theory}, pages 1858--1860,
  2003.

\bibitem[Freedman and Diaconis(1981)]{diaconis}
David Freedman and Persi Diaconis.
\newblock On the histogram as a density estimator: L2 theory.
\newblock \emph{Zeitschrift f{\"u}r Wahrscheinlichkeitstheorie und Verwandte
  Gebiete}, pages 453--476, 1981.

\bibitem[Goodfellow et~al.(2014)Goodfellow, Pouget-Abadie, Mirza, Xu,
  Warde-Farley, Ozair, Courville, and Bengio]{goodfellow2014generative}
Ian Goodfellow, Jean Pouget-Abadie, Mehdi Mirza, Bing Xu, David Warde-Farley,
  Sherjil Ozair, Aaron Courville, and Yoshua Bengio.
\newblock Generative adversarial nets.
\newblock In \emph{Advances in neural information processing systems}, pages
  2672--2680, 2014.

\bibitem[Gulrajani et~al.(2017)Gulrajani, Ahmed, Arjovsky, Dumoulin, and
  Courville]{gulrakani2017wassersteingp}
Ishaan Gulrajani, Faruk Ahmed, Martin Arjovsky, Vincent Dumoulin, and Aaron~C
  Courville.
\newblock Improved training of wasserstein {GAN}s.
\newblock In \emph{Advances in Neural Information Processing Systems}, pages
  5767--5777. 2017.

\bibitem[Heusel et~al.(2017)Heusel, Ramsauer, Unterthiner, Nessler, Klambauer,
  and Hochreiter]{heusel2017gans}
Martin Heusel, Hubert Ramsauer, Thomas Unterthiner, Bernhard Nessler,
  G{\"u}nter Klambauer, and Sepp Hochreiter.
\newblock {GAN}s trained by a two time-scale update rule converge to a {N}ash
  equilibrium.
\newblock \emph{arXiv preprint arXiv:1706.08500}, 2017.

\bibitem[Hindupur(2017)]{theganzoo}
Avinash Hindupur.
\newblock The {GAN} zoo.
\newblock \emph{GitHub Repository}, 2017.
\newblock URL \url{https://github.com/hindupuravinash/the-gan-zoo}.

\bibitem[Huang et~al.(2018)Huang, Yuan, Xu, Guo, Sun, Wu, and
  Weinberger]{huang2018an}
Gao Huang, Yang Yuan, Qiantong Xu, Chuan Guo, Yu~Sun, Felix Wu, and Kilian
  Weinberger.
\newblock An empirical study on evaluation metrics of generative adversarial
  networks.
\newblock 2018.
\newblock URL \url{https://openreview.net/forum?id=Sy1f0e-R-}.

\bibitem[Im et~al.(2018)Im, Ma, Taylor, and Branson]{divergencetraining}
Daniel~Jiwoong Im, Alllan~He Ma, Graham~W. Taylor, and Kristin Branson.
\newblock Quantitatively evaluating {GAN}s with divergences proposed for
  training.
\newblock In \emph{International Conference on Learning Representations}, 2018.
\newblock URL \url{https://openreview.net/forum?id=SJQHjzZ0-}.

\bibitem[Jolicoeur-Martineau(2018)]{ragan}
Alexia Jolicoeur-Martineau.
\newblock The relativistic discriminator: {A} key element missing from standard
  {GAN}.
\newblock \emph{arXiv preprint arXiv:1807.00734}, 2018.

\bibitem[Kingma and Ba(2015)]{adam}
Diederik Kingma and Jimmy Ba.
\newblock Adam: {A} method for stochastic optimization.
\newblock In \emph{International Conference on Learning Representations}, 2015.

\bibitem[Kodali et~al.(2017)Kodali, Abernethy, Hays, and
  Zsolt]{kodali2017dragan}
Naveen Kodali, Jacob Abernethy, James Hays, and Kira Zsolt.
\newblock On convergence and stability of {GAN}s.
\newblock \emph{arXiv preprint arXiv:1705.07215}, 2017.

\bibitem[Kullback and Leibler(1951)]{kullbackleibler}
Soloman Kullback and Richard Leibler.
\newblock On information and sufficiency.
\newblock pages 79--86, 1951.

\bibitem[Lee et~al.(2017)Lee, Ge, Ma, Risteski, and Arora]{lee2017ability}
Holden Lee, Rong Ge, Tengyu Ma, Andrej Risteski, and Sanjeev Arora.
\newblock On the ability of neural nets to express distributions.
\newblock In \emph{Journal of Machine Learning Research}, pages 1--26, 2017.

\bibitem[Lucic et~al.(2017)Lucic, Kurach, Michalski, Gelly, and
  Bousquet]{lucic2017gans}
Mario Lucic, Karol Kurach, Marcin Michalski, Sylvain Gelly, and Olivier
  Bousquet.
\newblock Are {GAN}s created equal? {A} large-scale study.
\newblock \emph{arXiv preprint arXiv:1711.10337}, 2017.

\bibitem[Mao et~al.(2017)Mao, Li, Xie, Lau, Wang, and Smolley]{lsgan}
Xudong Mao, Qing Li, Haoran Xie, Raymond~YK Lau, Zhen Wang, and Stephen~Paul
  Smolley.
\newblock Least squares generative adversarial networks.
\newblock In \emph{2017 IEEE International Conference on Computer Vision
  (ICCV)}, pages 2813--2821, 2017.

\bibitem[Mroueh and Sercu(2017)]{fishergan}
Youssef Mroueh and Tom Sercu.
\newblock Fisher {GAN}.
\newblock In \emph{Advances in Neural Information Processing Systems}, pages
  3310--3320, 2017.

\bibitem[Nowozin et~al.(2016)Nowozin, Cseke, and Tomioka]{fgan}
Sebastian Nowozin, Botond Cseke, and Ryota Tomioka.
\newblock f-{GAN}: {T}raining generative neural samplers using variational
  divergence minimization.
\newblock In \emph{Advances in Neural Information Processing Systems}, pages
  271--279. 2016.

\bibitem[Salimans et~al.(2016)Salimans, Goodfellow, Zaremba, Cheung, Radford,
  and Chen]{salimans2016improved}
Tim Salimans, Ian Goodfellow, Wojciech Zaremba, Vicki Cheung, Alec Radford, and
  Xi~Chen.
\newblock Improved techniques for training {GAN}s.
\newblock In \emph{Advances in Neural Information Processing Systems}, pages
  2234--2242, 2016.

\bibitem[Santurkar et~al.(2017)Santurkar, Schmidt, and Madry]{santurkar}
Shibani Santurkar, Ludwig Schmidt, and Aleksander Madry.
\newblock A classification-based perspective on {GAN} distributions.
\newblock \emph{arXiv preprint arXiv: 1711.00970}, 2017.

\bibitem[Theis et~al.(2015)Theis, van~den Oord, and Bethge]{theis}
Lucas Theis, Aäron van~den Oord, and Matthias Bethge.
\newblock A note on the evaluation of generative models.
\newblock \emph{arXiv preprint arXiv:1511.01844}, 2015.

\bibitem[Villani(2008)]{villani2008optimal}
C{\'e}dric Villani.
\newblock \emph{Optimal Transport: {O}ld and New}, volume 338.
\newblock Springer Science \& Business Media, 2008.

\end{thebibliography}
\end{document}


\title{Appendix for Generalizable Framework for Empirical Evaluation of Generative Adversarial Networks}
\author{Shayne O'Brien, Matt Groh, Abhimanyu Dubey\\
\{shayneob, groh, dubeya\}@mit.edu \\
Massachusetts Institute of Technology\\
}
\maketitle

\begin{enumerate}
\item Goodfellow et al 2014 (Generative adversarial networks, original assumptions)
\item Lucic et al 2017 (Are all GANs created equal)
\item Arora et al 2017a (Generalization and Equilibrium in Generative Adversarial Nets)
\item Arora et al 2017b (Do GANs actually learn the distribution? An empirical study) \cite{arora}
\item Santurkar et al 2018 (A Classification-Based Perspective on GAN Distributions) \cite{santurkar}
\item Li et al 2017 (Towards Understanding the Dynamics of Generative Adversarial Networks) \cite{li}
\end{enumerate}

In the following subsections, we outline their key differences in the following subsections.

\subsubsection{MMGAN} 
In the minimax GAN setup, the discriminator $D$ is trained to maximize the log probability of assigning the correct binary label to a given input for whether it is real or generated. The generator $G$ is trained to minimize the probability that $D$ will assign outputs from $G$ a low probability of being generated \cite{goodfellow2014generative}.

\subsubsection{NSGAN}
As stated by the authors, the MMGAN's generator struggles to learn well early in training because $D$ can reject samples from the randomly initialized $G$ with high confidence. This causes the objective of MMGAN to saturate if $G$ is not initialized properly.\footnote{MMGANs can successfully initialized by allowing $G$ to train for some number of steps, e.g. five, before beginning the simultaneous training regime.} To circumvent this, they propose training $G$ to maximize $\log(D(G(z))$ instead to provide stronger, non-saturating gradients in the initial training steps \cite{goodfellow2014generative}.

\subsubsection{WGAN}
As another alternative to the MMGAN, Arjovsky et al. \citeyear{arjovsky2017wasserstein} propose a modification to $D$ that allows it to output a scalar rather than a probability. They prove that, under this modification, minimizing the objective of $G$ with respect to a Lipschitz smooth $D$ corresponds to minimizing the Wasserstein distance between the real and generated distribution, where $D$ is a rough approximation to the otherwise intractable Wasserstein distance. The Lipschitz condition for $D$ is crudely enforced via weight clipping \cite{arjovsky2017wasserstein}.

\subsubsection{WGANGP}
Instead of relying upon weight clipping to enforce a Lipschitz constraint, the WGANGP introduces penalties to the loss as a function of the gradient's norm for improved WGAN training. The replacement of weight clipping with a gradient norm penalty makes it so the network does not suffer from the biasing of $D$ toward simple functions and the network does not require batch normalization, which implicitly changes the discriminator's task from one-to-one to many-to-many \cite{gulrakani2017wassersteingp}.

\subsubsection{DRAGAN}
Gradient penalties can further be introduced to the MMGAN and NSGAN by evaluating the penalty around the data manifold. The authors state that this encourages $D$ to be piecewise linear in these areas. Although the authors apply these penalties only to $D$, it is important to note that they can similarly be applied to $G$ \cite{kodali2017dragan}.

\subsubsection{LSGAN}
As opposed to other GANs that focus on minimizing KL-divergence or Wasserstein distance as their evaluation objective, the LSGAN to use a least-squares inspired loss as it implicitly minimizes the Pearson $\chi^{2}$ divergence metric. By optimizing according to this objective, the authors claim that the loss becomes smoother and saturates more slowly than that of the MMGAN \cite{mao2017least}.

\subsubsection{BEGAN}
Departing from the architecture laid out by Goodfellow et al. \citeyear{goodfellow2014generative}, the BEGAN uses an autoencoder as its $D$
and then optimizes a lower bound of the Wasserstein distance between the $D$ loss distributions on real 
and generated data. To control the equilibrium between $D$ and $G$, the authors introduce the hyperparameter $\gamma$. They cite $\gamma$ as necessary given that $D$ is tasked with autoencoding in addition to discriminating \cite{berthelot2017began}.

\section*{Data}
We evaluate the different GAN architectures using a synthetic data distribution framework.

\subsection*{Synthetic Data}
We generate multivariate distributions from the following families of distributions:

\subsubsection{Gaussian} The PDF of the multivariate Gaussian distribution follows:
\begin{equation}
{\displaystyle {\begin{aligned}p({\bf x})&={\frac {\exp \left(-{\frac {1}{2}}({\mathbf {x} }-{\boldsymbol {\mu }})^{\mathrm {T} }{\boldsymbol {\Sigma }}^{-1}({\mathbf {x} }-{\boldsymbol {\mu }})\right)}{\sqrt {(2\pi )^{k}|{\boldsymbol {\Sigma }}|}}}\end{aligned}}}
\end{equation}
where ${\bf x} \in \mathbb R^n$. The dimensionality $n$ is varied from 1 to 50, and for every experiment, the means ${\boldsymbol \mu}$ and covariances ${\bf \Sigma}$ are generated randomly from $[0,1]$ such that ${\bf \Sigma}$ is positive-definite.

\subsubsection{Exponential} The PDF of the multivariate exponential distribution follows:
\begin{equation}
p({\bf x}) = \prod_{i=1}^n \left( \lambda_i \right) \exp(-{\boldsymbol \lambda}^\top {\bf x}) \ \forall x \geq 0
\end{equation}
where ${\bf x} \in \mathbb R^n$. The dimensionality $n$ is varied from 1 to 50, and for every experiment, we generate the inverse mean vector ${\boldsymbol \lambda} = \{\lambda_1, ..., \lambda_n\}$ is generated randomly from $[0,1]$.

\subsubsection{Beta} The PDF of the multivariate Beta (Dirichlet) distribution follows:
\begin{equation}
p({\bf x}) = \frac{\Gamma(\sum_{i=1}^n \alpha_i)}{\prod_{i=1}^n \Gamma(\alpha_i)} x_i^{\alpha_i}
\end{equation}
where ${\bf x} = (x_1, ..., x_n) \in \mathbb (0,1)^n$ such that $\sum_{i=1}^n x_i = 1$. The unnormalized mean vector ${\boldsymbol \alpha}$ is chosen randomly from $[0,1]$.

\subsubsection{Gamma} The PDF of the multivariate Gamma distribution is given by:
\begin{equation}
p({\bf x} = \prod_{i=1}^n \frac{1}{\Gamma(k_i)\theta_i^{k_i}}x_i^{k_i - 1}\exp(-\frac{x}{\theta_i})
\end{equation}
The parameters $k_i$ and $\theta_i$ are chosen randomly from the ranges $[1,1000]$ and $(0,1)$ respectively.

\subsubsection{Gumbel} The PDF of the multivariate Gumbel distribution is given by:
\begin{equation}
p({\bf x}) = \prod_{i=1}^n \left(\exp\left( - \exp\left( -\frac{x_i - \mu_i}{\beta_i}\right)\right)\right)
\end{equation}
where ${\bf x} \in \mathbb R^n$. The dimensionality $n$ is varied from 1 to 50, and for each experiment the vectors ${\boldsymbol \mu} = \{\mu_1, ..., \mu_n\}$ and ${\boldsymbol \beta} = \{\beta_1, ..., \beta_n\}$, are chosen from $[0,1]$.

\subsubsection{Laplace distributions} The PDF of the multivariate Laplace distribution is given by:
\begin{gather}
p({\bf x}) = \frac{1}{2^n} \prod_{i=1}^n \frac{1}{b_i} \exp\left(-\frac{|x - \mu_i |}{b_i}\right)
\end{gather}
The dimensionality $n$ is varied from 1 to 50, and for every experiment, the means ${\boldsymbol \mu}$ and half-variances ${\bf b} = \{b_1, ..., b_n\}$ are chosen randomly from $[0,1]$.

\subsection*{Mixture Models}
To evaluate the extent of \textit{mode collapse}, we generate mixture probabilities with the following rule:
\begin{equation}
p({\bf x}) = \frac{1}{m}\sum_{i=1}^m p_i({\bf x})
\end{equation}
where $p_i$ is drawn from the above mentioned probability distributions. 

In our study, we restrict ourselves to equally-likely modes and hence set the mixture weights as uniform. Moreover, all $p_i$ are taken from the same family of distributions. This reduces the overall complexity of the mixture probability distribution as we observed that non-uniform mixing and mixing different distributions proved to be too complicated for our simple feedforward neural network GAN architecture.

\subsubsection{KL-Divergence} The most common metric used in multiclasss machine learning, we calculate the KL-divergence of $p_g$ from $p_{data}$. Under the independence assumption, however, we can simplify this further:
\begin{equation}
\mathbb D_{\sf KL}[p_{data} || p_g] = \sum_{i=1}^n \mathbb D_{\sf KL}[p_{data}^{(i)} || p_{g}^{(i)}]
\end{equation}
where $\mathbb D_{\sf KL}(P || Q) = \int_{t=-\infty}^\infty p(t) \log(\frac{p(t)}{q(t)}) \geq 0$. This implies that we can compute each of the metrics dimension-wise using marginals instead of computing the divergence of the joint, which will be difficult since the histograms would be sparse in that case \cite{kullback}.

\subsubsection{Jensen-Shannon Divergence} The JS Divergence is a symmetrized and normalized version of the KL-divergence. Here as well, with the independence assumption, the computation is simplified to the sum of dimension-wise elements. The JS Divergence is given by:
\begin{equation}
\small
\mathbb D_{\sf JS}[P || Q] = \frac{1}{2}\left[\mathbb D_{\sf KL}[P || \frac{1}{2}(P + Q)] + \mathbb D_{\sf KL}[Q || \frac{1}{2}(P + Q)]\right]
\end{equation}
where $P$ and $Q$ are probability distributions \cite{jensenshannon}. 

\subsubsection{Wasserstein Distance} The Wasserstein Distance has become the metric of choice for training GANs. Typically computing the Wasserstein Distance is intractable and approximations are frequently used when using it as an optimization criterion~\cite{arjovsky2017wasserstein}. In our case, we compute the exact Wasserstein difference between the two histogram approximations under the independence assumption, and hence it becomes the sum of dimension-wise Wasserstein distances. The Wasserstein distance is given by:
\begin{equation}
l_1(u, v) = \int_{-\infty}^\infty |U - V|
\end{equation}
where $U$ and $V$ are cumulative distribution functions \cite{villani2008optimal}.

\subsubsection{Energy Distance} The Energy Distance is a statistical distance that is calculated directly from the samples themselves. This makes it easy measure to compute without having to make approximations for $p_g$ and $p_{data}$. It is given by:
\begin{equation}
D^2_{\sf EN}(P, Q) = 2\mathbb E|| X - Y|| - \mathbb E|| X - X'|| - \mathbb E||Y - Y'|| \geq 0
\end{equation}
where $X, X' \sim P$ and $Y, Y' \sim Q$. While this metric has not appeared popularly in machine learning contexts, it is easy to compute and uniformly converges to the true energy distance in the presence of large number of samples~\cite{szekely2013energy}.

\begin{table*}[tbp!]
\centering
\begin{tabular}{l|c|c|c|c|c|c}
\hline
\hline
\multirow{2}{*}{GAN Model} & \multicolumn{6}{l}{Average Log-likelihood of Multivariate Normal (n=5, 10000 samples)} \\ \cline{2-7}
                          & N=10          & N=15          & N=25          & N=50          & N=75         & N=100        \\ \hline
True Data                  & -1100.21      & -1147.23      & -1109.42      & -1203.33      & -1225.32     & -1098.34     \\
MMGAN                      & -2868.61      & -2700.95      & -2359.69      & -2000.94      & -1668.73     & -1440.27     \\
NSGAN                      & -2774.93      & -2684.87      & -2589.45      & -2304.88      & -2197.45     & -1983.53     \\
WGAN                       & -2765.23      & -2733.83      & -2372.28      & -1977.48      & -1609.34     & -1339.89     \\
WGANGP                     & -2669.22      & -2505.43      & -2288.31      & -1901.39      & -1708.29     & -1421.18     \\
DRAGAN                     & -2889.87      & -2680.76      & -2538.92      & -2237.29      & -2003.03     & -1880.38     \\
LSGAN                      & -2780.98      & -2590.94      & -2353.55      & -2104.96      & -1804.48     & -1659.44     \\
BEGAN                      & -2520.41      & -2449.23      & -2375.82      & -2093.85      & -1889.39     & -1807.76   \\ \hline
\end{tabular}
\label{tab:redundancy}
\caption{The log-likelihood of the generated data under the true distribution increases as the redundancy in the data is increased.}
\end{table*}

\subsection{Hyperparameter Tuning}

In the original GAN papers, each GAN has a suggested set of default hyperparameters, which we test for each model. In practice, however, the hyperparameters for the optimal GAN differ depending on the dataset, so we also conduct a computationally reasonable hyperparameter search that takes less than 48 hours on a single, mid-range GPU. Our hyperparameter search focuses on three hyperparameters: the learning rates of the generator and discriminator (.001, .0005, .0001, .00005), the number of hidden dimensions (16, 32, 64, 128, 256), and the batch size (100, 150, 200, 250). We ran the hyperparameter experiment on data distributions with 100 dimensions and 10,000 observations. Each GAN was trained for 10 epochs without early stopping. 

Due to our decision to keep the step size as the default indicated by the original GAN papers, we report results with varying step counts across GANs. We optimize the hyperparameters on the generator loss. Table~\ref{table:t1} presents results of the hyperparameter tuning that demonstrate hyperparameters are rarely shared across the seven GANs for the same data distribution.

In case of the multivariate Gaussian distributions and data with redundancy, we first explored the estimation of $p_g$ using Kernel Density Estimation (KDE), however, we found that this was computationally expensive and its observed divergence metrics had similar trends to the histogram estimation cases across models. In the interest of simplicity and reproducibility, we adopted the histogram estimation technique.